\documentclass{article}

\usepackage[preprint]{neurips_2024}

\usepackage[utf8]{inputenc} % allow utf-8 input
\usepackage[T1]{fontenc}    % use 8-bit T1 fonts
\usepackage{hyperref}       % hyperlinks
\usepackage{graphicx}
\usepackage{amsmath}
\usepackage{url}            % simple URL typesetting
\usepackage{booktabs}       % professional-quality tables
\usepackage{amsfonts}       % blackboard math symbols
\usepackage{nicefrac}       % compact symbols for 1/2, etc.
\usepackage{microtype}      % microtypography
\usepackage{subfig} 
\usepackage{xcolor}         % colors

\title{Virtual avatar generation models as world navigators}

\author{%
  Sai Mandava\\
  SABR\\
  \texttt{saimandava8@gmail.com} \\
}

\begin{document}

\maketitle

\begin{figure}[ht]
  \centering
  \includegraphics[height=3cm]{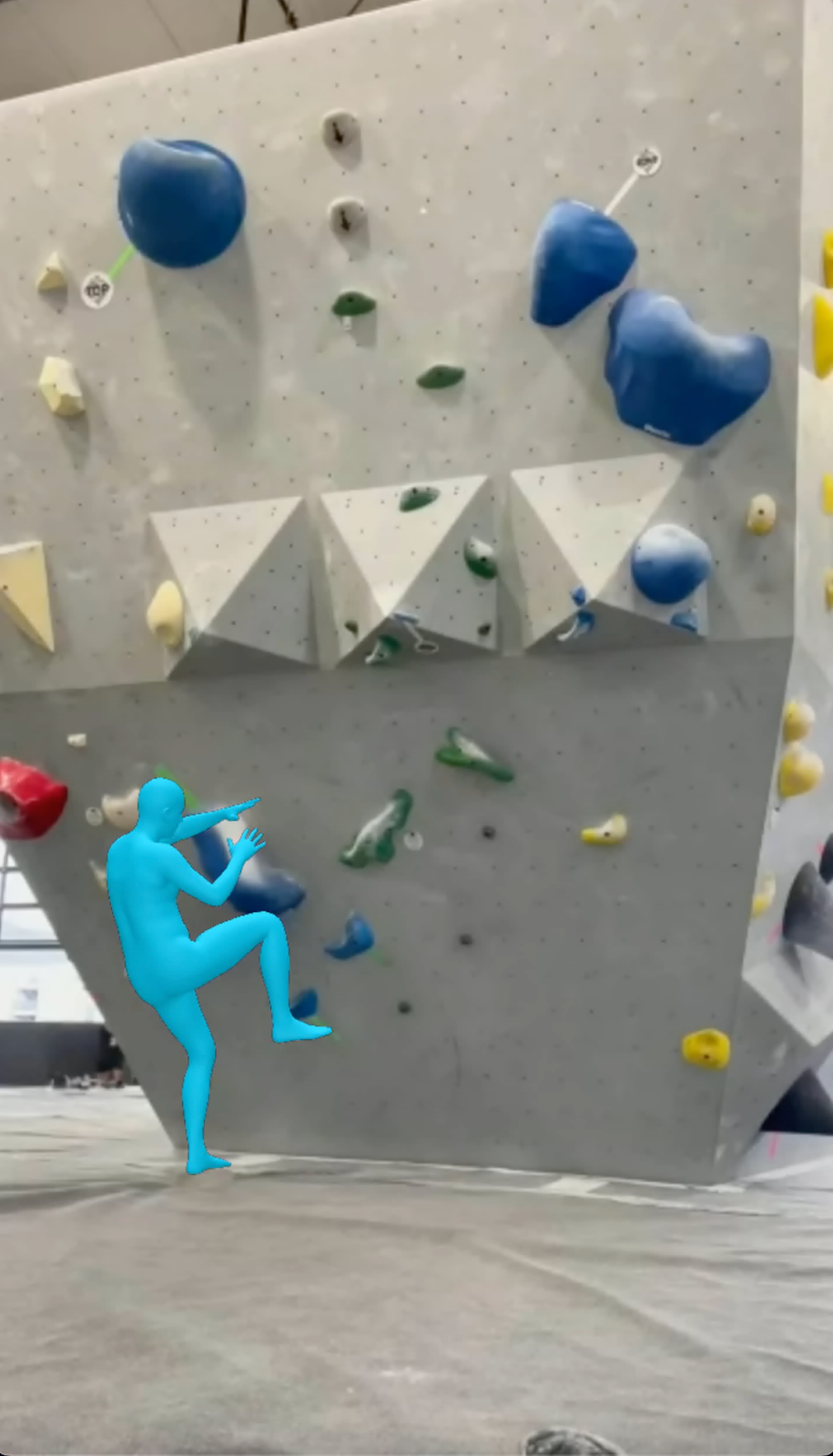}
  \includegraphics[height=3cm]{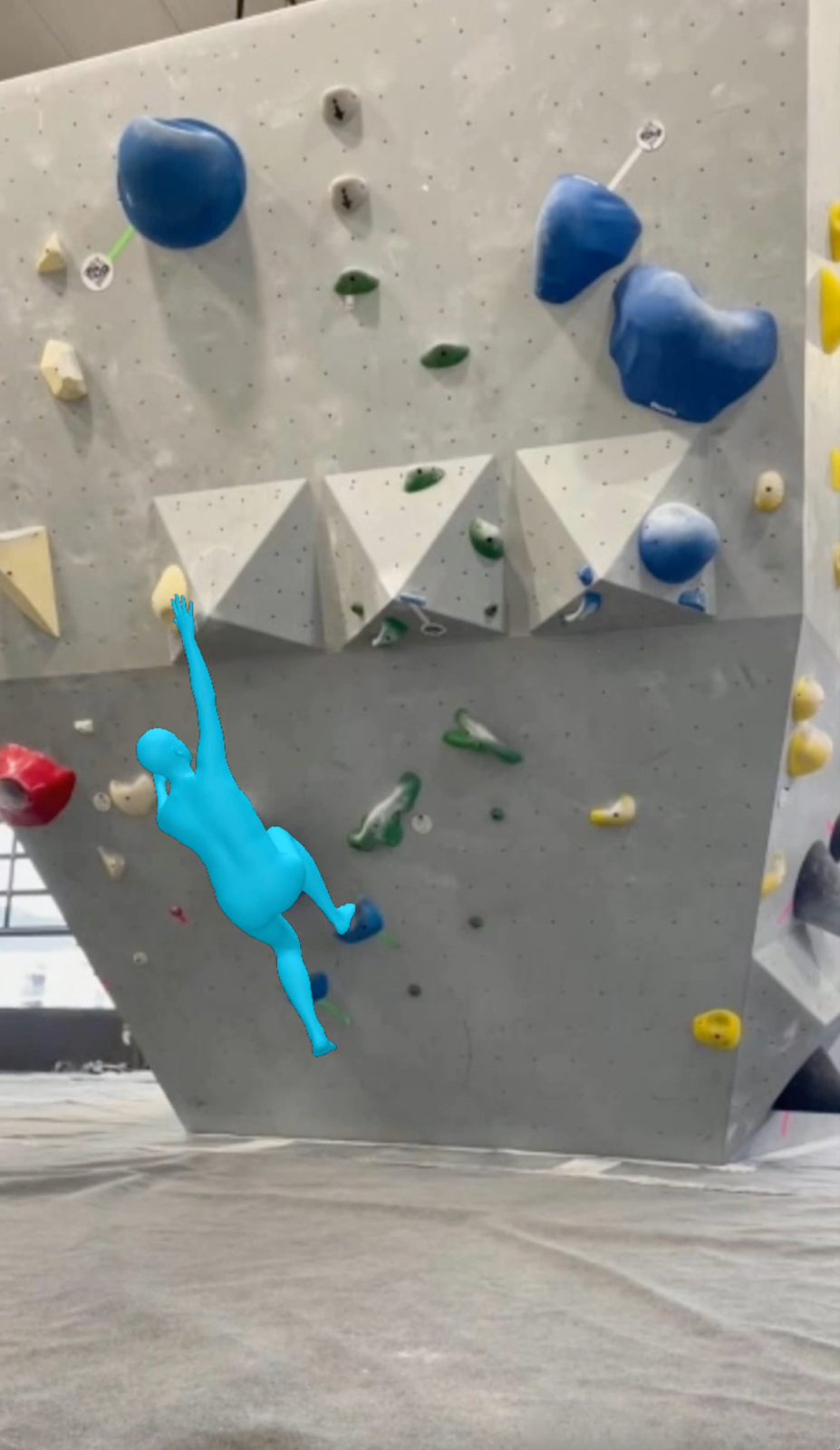}
  \includegraphics[height=3cm]{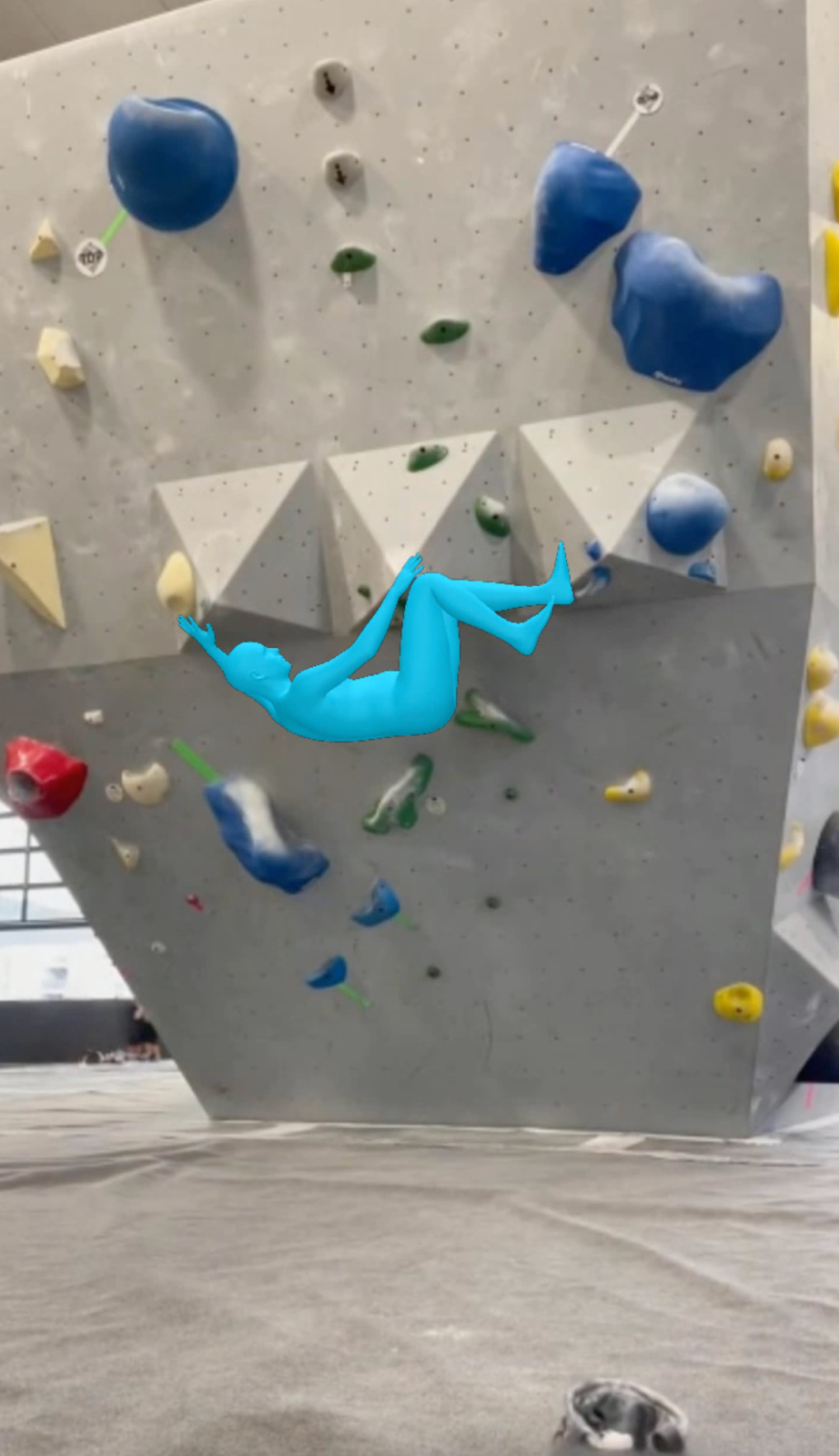}
  \includegraphics[height=3cm]{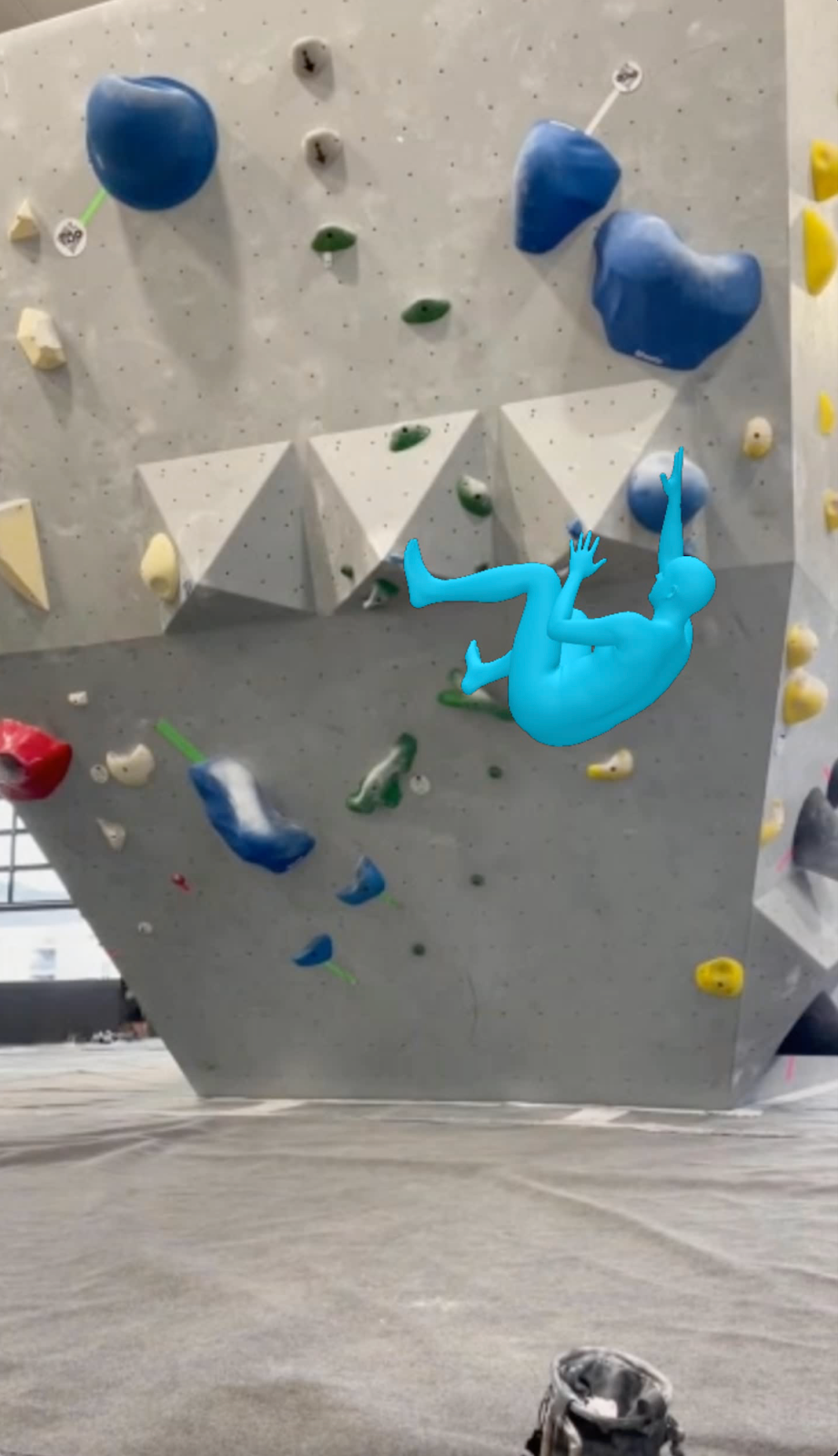}
  \includegraphics[height=3cm]{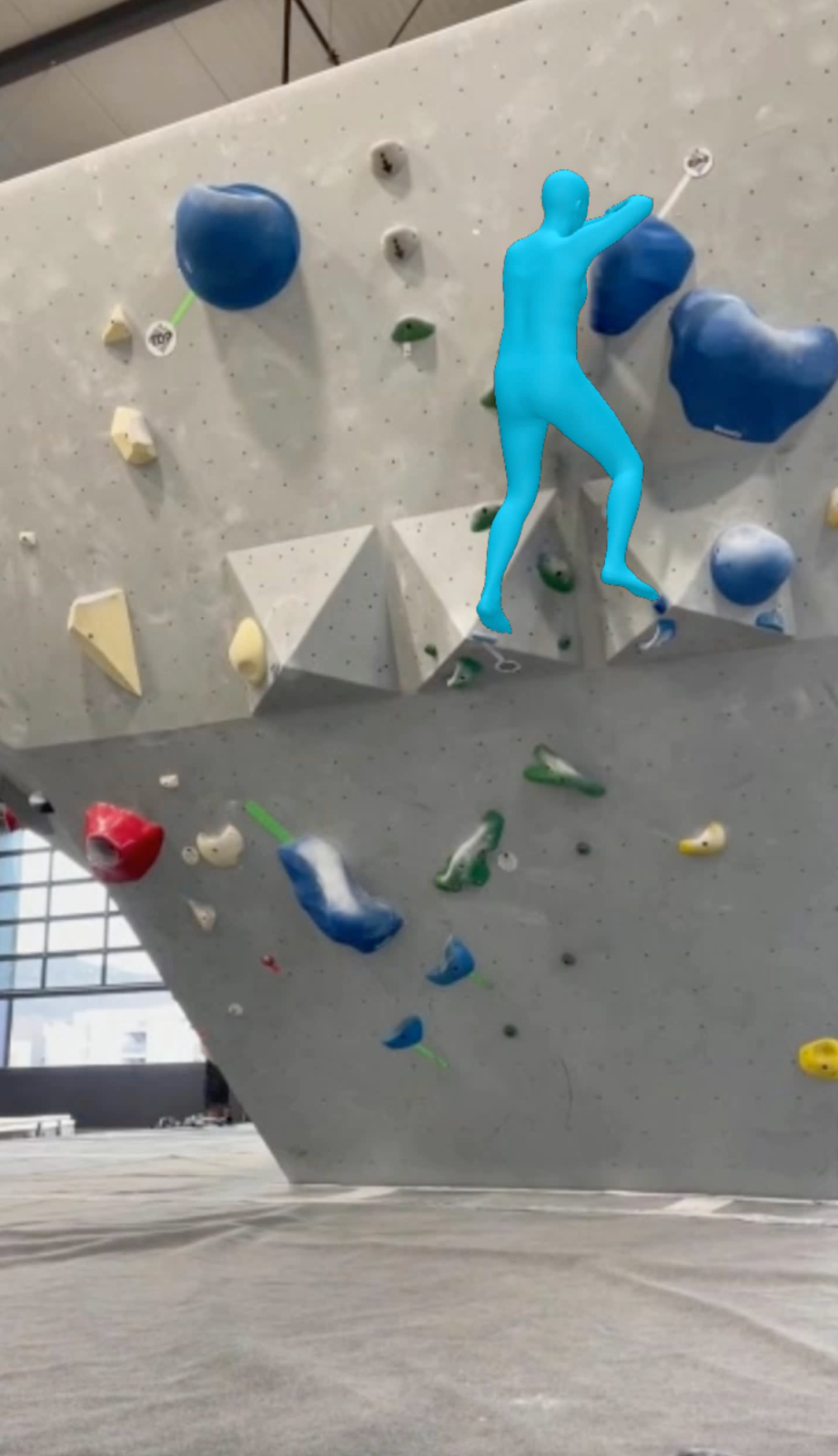}
  \caption{Indoor, blue route}
  \label{fig:yourlabel1}
\end{figure}

\begin{figure}[ht]
  \centering
  \includegraphics[height=3cm]{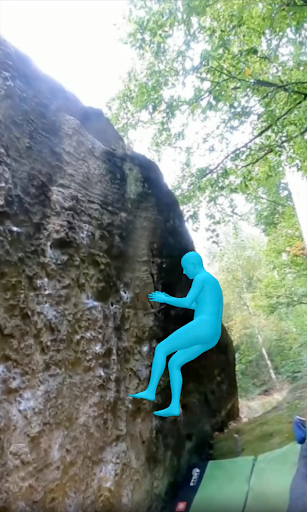}
  \includegraphics[height=3cm]{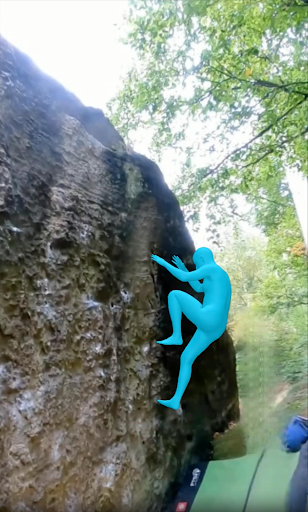}
  \includegraphics[height=3cm]{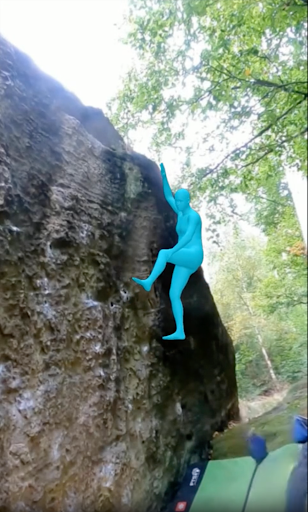}
  \includegraphics[height=3cm]{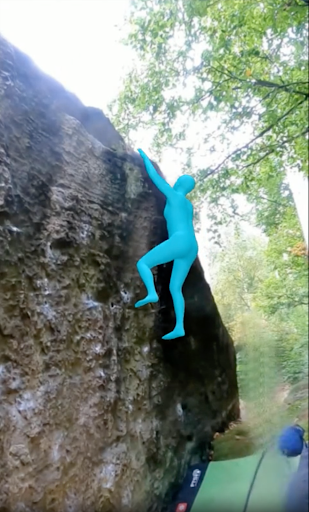}
  \includegraphics[height=3cm]{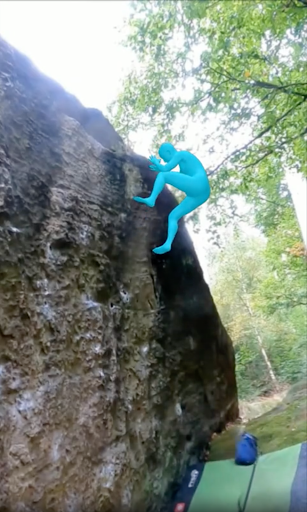}
  \caption{Outdoor route}
  \label{fig:yourlabel3}
\end{figure}

\begin{abstract}
  We present a novel video model capable of simulating human movement within a given environment by assuming the parameters of a virtual avatar. Our model, a diffusion transformer applied to the human motion domain, features two key design choices: predicting the sample instead of noise in each diffusion step, and ingesting entire videos to predict complete motion sequences. We focus on rock climbing environments because they are controlled, single-agent, and static, yet encompass many complex biomechanical interactions found in various real-world scenarios. We aptly name our model SABR-CLIMB. Analogous to text-to-video models [\hyperlink{ref37}{37}, \hyperlink{ref39}{39}], we show that with sufficient data and computational power, diffusion models [\hyperlink{ref7}{7}, \hyperlink{ref8}{8}, \hyperlink{ref18}{18}] can implicitly model 3D relationships—in our case, between human biomechanics and the 3D world, using 2D RGB videos as input. The last decade of Machine Learning research has demonstrated that rich data, simple learning algorithms, and powerful computational resources outperform hand-engineered systems in computer vision, translation, and NLP \hyperlink{ref38}{[38]}. We apply this framework to a new problem, compiling a large-scale, proprietary dataset of 22 million environment sequences and corresponding individual motion tracks. We call this dataset NAV-22M. Our results indicate that training a virtual avatar generation model on real-world environment videos, leveraging substantial computational resources, and enhancing dataset size and quality are promising strategies for developing general-purpose virtual avatars capable of navigating the world from a human perspective without any reward design or skill primitives. Our full codebase and example videos can be found on our project page - \href{https://virtual-avatar-generation.github.io/}{https://virtual-avatar-generation.github.io/}.
\end{abstract}

\section{Introduction}

Human motion generation has numerous applications, from computer animation to robotics. However, challenges in data acquisition and the complexity of modeling diverse motions have limited research in this field. Most existing models in this field focus on the text-to-motion setting [\hyperlink{ref1}{1}, \hyperlink{ref4}{4}, \hyperlink{ref5}{5}, \hyperlink{ref9}{9}, \hyperlink{ref10}{10}, \hyperlink{ref11}{11}, \hyperlink{ref12}{12}, \hyperlink{ref42}{42}, \hyperlink{ref44}{44}, \hyperlink{ref45}{45}] while some explore the action-to-motion setting \hyperlink{ref3}{[3]}, beneficial for computer animation but less so in other industries such as sports, biotech, healthcare, and robotics. In this paper, we investigate the video-to-motion setting. To meet user experience constraints, we use in-the-wild videos, typically recorded with modern phone cameras. Human motion generation faces the challenge of many-to-many mapping, with virtually infinite ways to move. This complexity is amplified in the video domain. Previous approaches using auto-encoders or VAEs \hyperlink{ref2}{[2]} imply one-to-one mapping or a normal latent distribution. Diffusion models [18, 54], however, are more suitable as they do not assume a target distribution and excel in many-to-many distribution matching. U-Net \hyperlink{ref17}{[17]} has been a common backbone for diffusion models, but recent architectures like DiT \hyperlink{ref50}{[50]} and U-ViT \hyperlink{ref51}{[51]}, which adapt ViT \hyperlink{ref52}{[52]} for diffusion models, show great performance in image generation. Attention-based architectures \hyperlink{ref19}{[19]}, which capture long-range contextual relationships, also offer a promising approach for videos. Extending DiT, we use a video transformer as the backbone, incorporating both spatial and temporal blocks \hyperlink{ref49}{[49]}. Our model processes entire environment videos (up to 45 seconds) to generate complete motion sequences. We chose rock climbing to showcase our model's capabilities because of its static environment and solo nature, making evaluation clearer.

In both computer vision and natural language processing, large pre-trained models are increasingly popular due to their broad downstream applications, which justify their scaling. Our video-to-motion model is such a large pre-trained model which could potentially be useful not just in computer vision, but also in robotics [\hyperlink{ref71}{71}, \hyperlink{ref72}{72}, \hyperlink{ref73}{73}], computer graphics \hyperlink{ref74}{[74]}, biomechanics \hyperlink{ref75}{[75]}, and other fields where simulations of virtual avatars are needed.

Our contributions can be summarized as follows: 

\begin{enumerate}
    \item We introduce a new type of foundation model, the virtual avatar generation model, designed for the video-to-motion setting. Here, the input is videos of empty environments, and the output is a sequence of suggested poses, representing the motion, within the video frame space.
    
    \item We propose a sequential pipeline architecture guided by the sequence of objects and materials the user interacts with, demonstrating its effectiveness in the rock climbing environment as a proof of concept. We aptly name our model SABR-CLIMB.
    
    \item We qualitatively demonstrate the model's effectiveness by showcasing multiple examples of the AI avatar climbing various indoor and outdoor walls, captured in different videography styles. We also quantitatively measure the model's success via two errors: route trajectory adherence and movement adherence. Our results establish SABR-CLIMB as the first ever, state-of-the-art model for virtual avatar navigation in this domain.
\end{enumerate}

\section{Related Work}

\subsection{Human Motion Generation}

Neural motion generation, learned from motion capture data, can be conditioned on various signals representing motion. Historically, these signals include text and motion segments. Some models predict motion from initial poses [\hyperlink{ref5}{5}, \hyperlink{ref14}{14}, \hyperlink{ref41}{41}, \hyperlink{ref59}{59}], while others [\hyperlink{ref40}{40}, \hyperlink{ref67}{67}, \hyperlink{ref68}{68}, \hyperlink{ref69}{69}] focus on in-betweening—filling in the middle sequence of poses given the start and end poses—using architectures like bi-directional GRU [\hyperlink{ref43}{43}] and Transformer [\hyperlink{ref19}{19}]. Additionally, some models [\hyperlink{ref44}{44}] use an auto-encoder to learn motion latent representation, enabling motion editing and control with spatial constraints such as root trajectory and bone lengths. Motion control can also be guided by high-level cues from action classes [\hyperlink{ref3}{3}, \hyperlink{ref45}{45}, \hyperlink{ref70}{70}], audio [\hyperlink{ref46}{46}, \hyperlink{ref47}{47}], and natural language [\hyperlink{ref9}{9}, \hyperlink{ref10}{10}]. Typically, specialized approaches are developed to map each conditioning domain into motion.

Our work builds on these methods by leveraging the strengths of neural motion generation while introducing a novel approach that integrates video-based conditioning. Similar to prior approaches, our model utilizes a well-structured latent space for robust motion priors. However, we differ by seamlessly integrating this latent space with our video encoder, allowing for direct conditioning on video inputs, which is less common in traditional motion generation models.

\subsection{Denoising diffusion probabilistic models (DDPMs)}

Diffusion models are neural generative models based on the stochastic diffusion process from thermodynamics. In these models, a sample from the data distribution is gradually noised through a diffusion process, and a neural model learns the reverse process of denoising the sample. Sampling from the learned data distribution involves denoising pure initial noise. While some diffusion models have been used for motion generation, most previous examples utilize text or image inputs rather than video.

In contrast, our work applies diffusion models to video-to-motion generation. We implement video-to-motion diffusion by conditioning on the full set of a pre-trained, frozen video encoder's patch tokens for all input video frames. This approach allows us to leverage the rich temporal information contained in videos, differentiating our method from traditional diffusion models that do not typically integrate video data as a conditioning input. Additionally, we incorporate common techniques from adjacent areas such as language modeling, like prompting, to guide motion generation. In our case, the prepended "prompt" consists of context information about which objects users interact with which ultimately will influence the manner of motion in the output video. By integrating these strategies, we enhance the capability of the diffusion model to generate coherent and contextually accurate motion sequences from video inputs. 

\section{Data Processing}
We assemble our curated NAV-22M dataset from various public and private sources of climbing videos. Below, we describe the main components of our comprehensive data pipeline, including data sources, deduplication, sanitation, and the dual sub-pipeline system to prepare inputs for our core model.

\begin{figure}[ht]
  \centering
  \includegraphics[height=3cm]{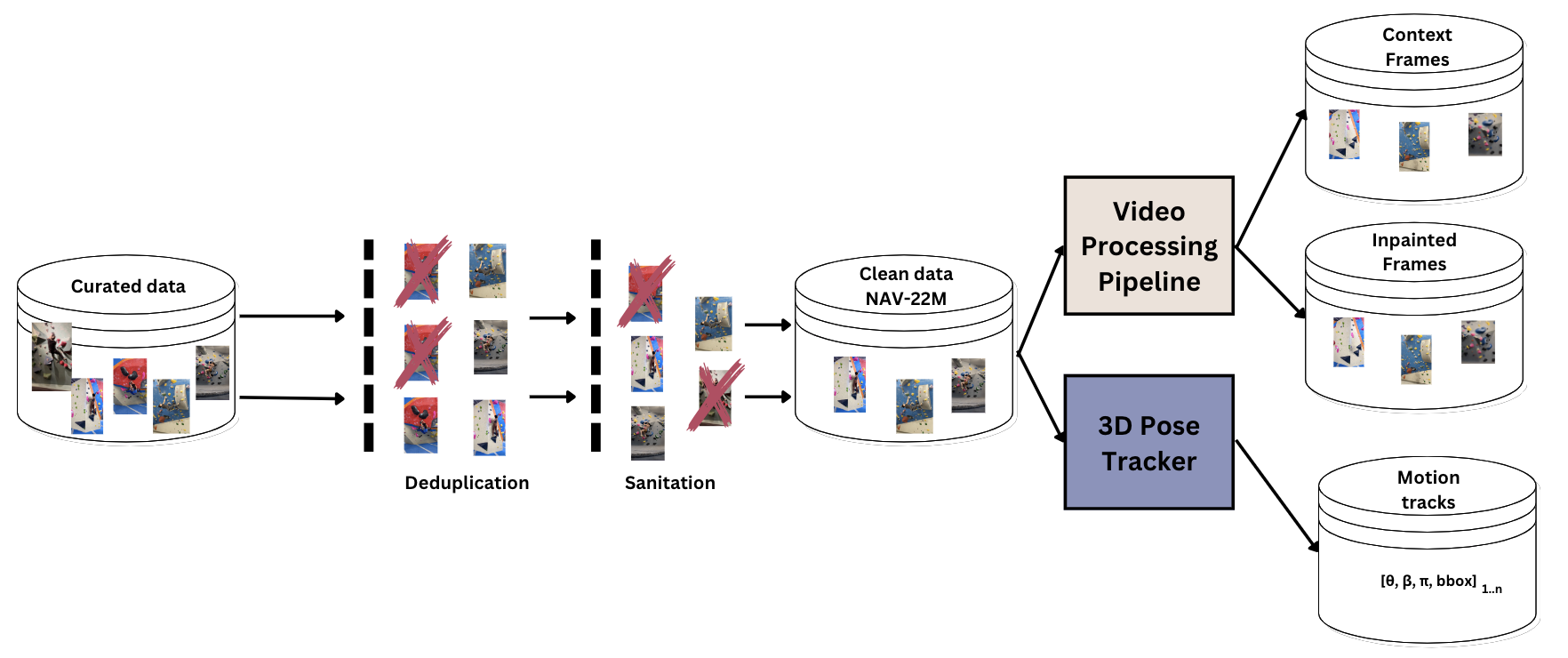}
  \hspace{0.3cm} % Add horizontal space between the images
  \vrule{} % Add vertical line
  \hspace{0.3cm} % Add horizontal space between the line and the next image
  \includegraphics[height=3cm]{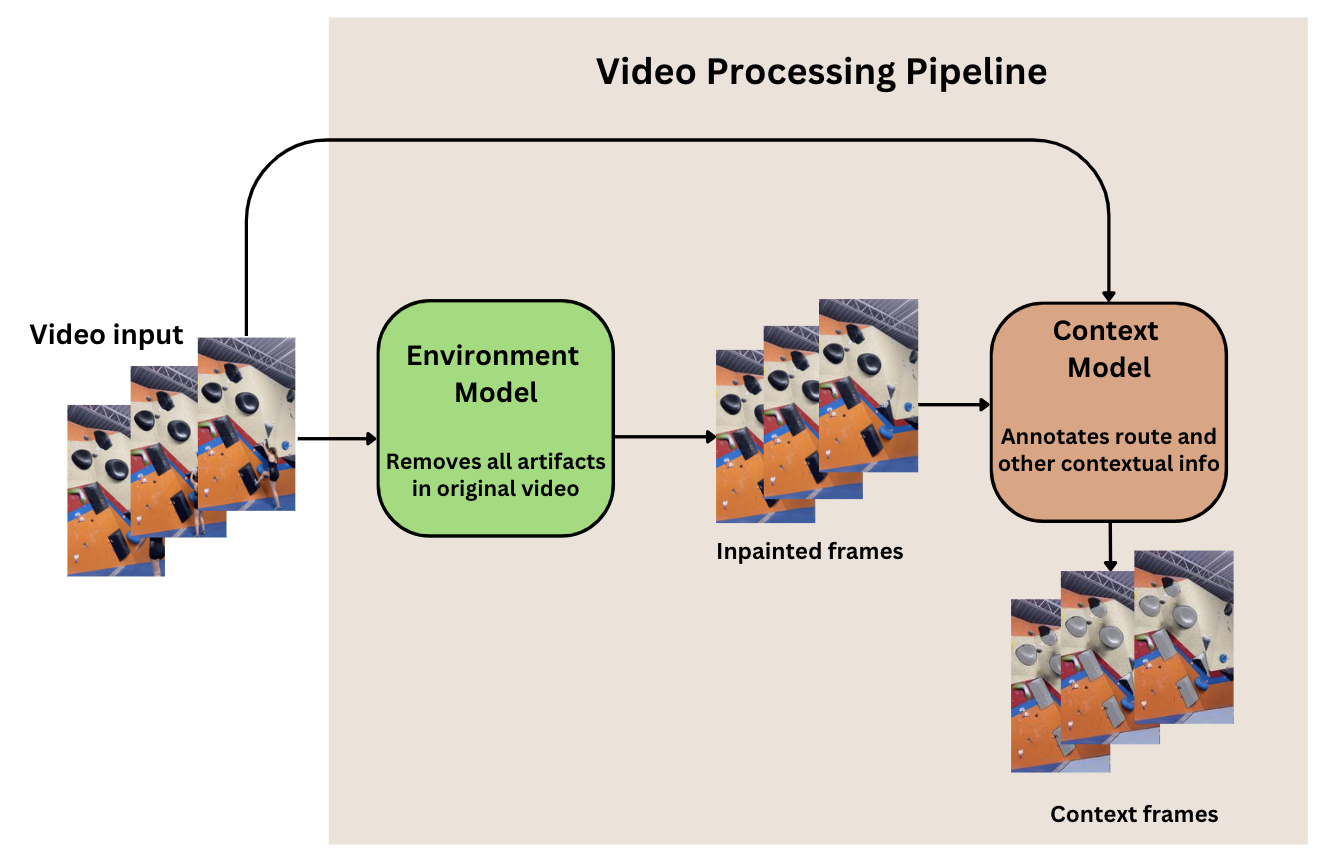}
  \caption{Overview of our data pipeline. Left: The data processing pipeline used to curate and prepare the NAV-22M dataset for training the SABR-CLIMB model. Right: Enhanced view of the the video processing pipeline.}
\end{figure}

\subsection{Data Sources}

A significant challenge is the limited availability of general video data, particularly for specific sports like rock climbing. To address this, we curated an extensive collection of proprietary climbing video data. Each video undergoes rigorous inspection and cleaning to remove motion and video artifacts, ensuring high-quality content. As shown in Figure 3, our process involves deduplication and sanitation phases. We then run a video processing pipeline and a slightly modified version of the off-the-shelf 3D pose tracker, PHALP \hyperlink{ref20}{[20]}, over each video. Each video in our dataset averages between 20 seconds to 1 minute, with accurate corresponding motion data. The motion data from PHALP captures the correct techniques for climbing specific environments, broadly encompassing all modern climbing techniques. We downsample all videos to a fixed resolution of 368x640 at 24 FPS, allowing our model to learn standardized motion sequences over a consistent number of patch tokens per video frame. This results in 22 million unique sequences for training our model.

\subsection{Deduplication}

We manually review each video in the curated dataset to remove duplicates, reducing redundancy and increasing diversity among the videos.

\subsection{Sanitation}

Our sanitation step involves cropping each video to capture the full movement between the start and end of a climb. We remove videos with artifacts (e.g., objects flying in front of the camera) or poor videography (e.g., extreme swinging or zooming of the camera).

\subsection{Video Processing Pipeline}
\begin{itemize}
    \item \textbf{Environment Model:} We track the main climber throughout the input video and apply a modified, long-context version of an off-the-shelf video inpainting model, FGT [\hyperlink{ref25}{25}] to remove all climber-related artifacts (and any other people) across the full video.
    \item \textbf{Context Model:} We segment [\hyperlink{ref29}{29}] and track the objects [\hyperlink{ref30}{30}] the main climber interacts with and prepend the unique frames of these segmented objects (signifying the "route" the person climbs) to the original clean video frames.
\end{itemize}

\subsection{3D Pose Tracker}

We use a slightly modified version of PHALP, an off-the-shelf tracker that predicts 3D pose parameters for each video frame. PHALP captures detailed 3D motion data, enabling us to accurately represent climbing techniques. We remove video frames with missing tracks to ensure continuous motion representation. This results in clean, high-quality videos with comprehensive motion data, suitable for training our model.

\section{Avatar Navigation}
\subsection{Preliminaries}
\paragraph{Body Model.}

The SMPL [\hyperlink{ref21}{21}] model is a low-dimensional parametric representation of the human body. It generates a 3D mesh \(M \in \mathbb{R}^{3 \times N}\) with \(N = 6890\) vertices based on input parameters for pose \(\theta \in \mathbb{R}^{24 \times 3 \times 3}\) and shape \(\beta \in \mathbb{R}^{10}\). Body joints \(X \in \mathbb{R}^{3 \times k}\) are computed as a linear combination of the mesh vertices, expressed as \(X = MW\), where \(W \in \mathbb{R}^{N \times k}\) are fixed weights. Pose parameters \(\theta\) include body pose parameters \(\theta_b \in \mathbb{R}^{23 \times 3 \times 3}\) and global orientation \(\theta_g \in \mathbb{R}^{3 \times 3}\).

\paragraph{Camera.}

We utilize a perspective camera model with fixed focal length and intrinsics \(K\) [\hyperlink{ref21}{21}]. Each camera \(\pi = (R, t)\) is characterized by a global orientation \(R \in \mathbb{R}^{3 \times 3}\) and translation \(t \in \mathbb{R}\). Given these parameters, points in the SMPL space (such as joints \(X\)) are projected onto the image plane as \(x = \pi(X) = \Pi(K(RX + t))\), where \(\Pi\) represents perspective projection using camera intrinsics \(K\). Since the pose parameters \(\theta\) already include global orientation, we typically assume \(R\) to be the identity matrix.

\paragraph{SABR-CLIMB.} The goal of the virtual avatar generation task is to learn a predictor \(f(V)\) that, given a video \(V\), estimates the 3D pose and shape of the person (avatar) in each frame. Following typical parametric approaches for human mesh recovery [\hyperlink{ref76}{76}, \hyperlink{ref77}{77}], we model \(f\) to predict \(\Theta_t = [\theta_t, \beta_t, \pi_t, b_t]\) for each video frame \(t\). Here, \(\theta_t\) and \(\beta_t\) are the SMPL pose and shape parameters, and \(\pi_t\) is the camera translation. Unlike standard human mesh recovery models, which estimate the 3D information of a human in an image, we also include the bounding box \(b_t = [x_t, y_t, w_t, h_t]\). This allows us to predict where the 3D human should be in each frame, rather than where it is. Our diffusion model processes a video sequence of up to 45 seconds (\(\sim\)1100 video frames), predicting a sequence of these parameters. The parameters \(\theta_t\) and \(\beta_t\) are used to create the 3D avatar mesh for each frame. The bounding box \(b_t\) and 3D camera translation \(\pi_t\) are combined to estimate the perspective camera translation via our weak perspective camera model, which is then used to visualize the 3D avatar mesh onto the respective video frame.

\subsection{Model Architecture}
\begin{figure}[ht]
  \centering
  \includegraphics[height=4cm]{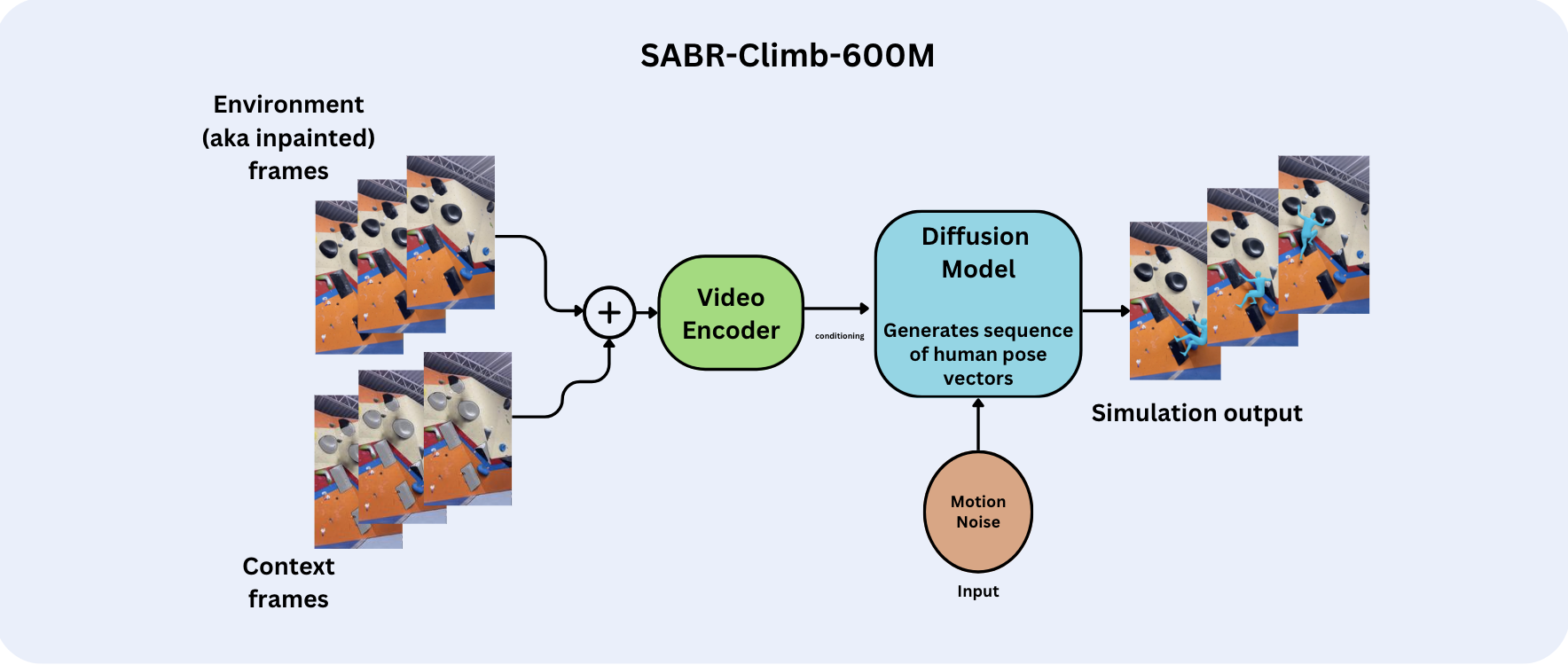}
  \caption{Overview of the SABR-Climb-600M model.}
\end{figure}

The training of SABR-CLIMB, our state-of-the-art virtual avatar generation foundation model, emphasizes integrating coarse and fine video details with human motion. To effectively process video data and create simulated videos, it is essential to analyze each frame and manage extended video lengths. As shown in Figure 4, SABR-CLIMB is an end-to-end diffusion model that is designed to take in environment videos and context frames for suggested motion, encode the full video with context, and predict the final motion sequence. This final motion sequence is then visualized across the input video frames.

\paragraph{Video Encoder.}

We utilize the state-of-the-art DINOv2 [\hyperlink{ref24}{24}] large model with registers [\hyperlink{ref23}{23}] as a frozen backbone for its exceptional embeddings. Registers are used because they lead to smoother feature maps and attention maps, enhancing downstream visual processing. Since DINOv2 has a patch size of 14, we downsample each frame further to 280x490 resolution to preserve the aspect ratio of the input video while also accommodating the patch size and maintaining computational efficiency. DINOv2 then processes each frame of the combined context and clean environment video frames into patch tokens, which are then fed as conditioning into the diffusion model. DINOv2 embeddings have proven to significantly encapsulate fine-grained details, coarse-grained details, and relevant depth information.

\begin{figure}[ht]
  \centering
  \includegraphics[height=4cm]{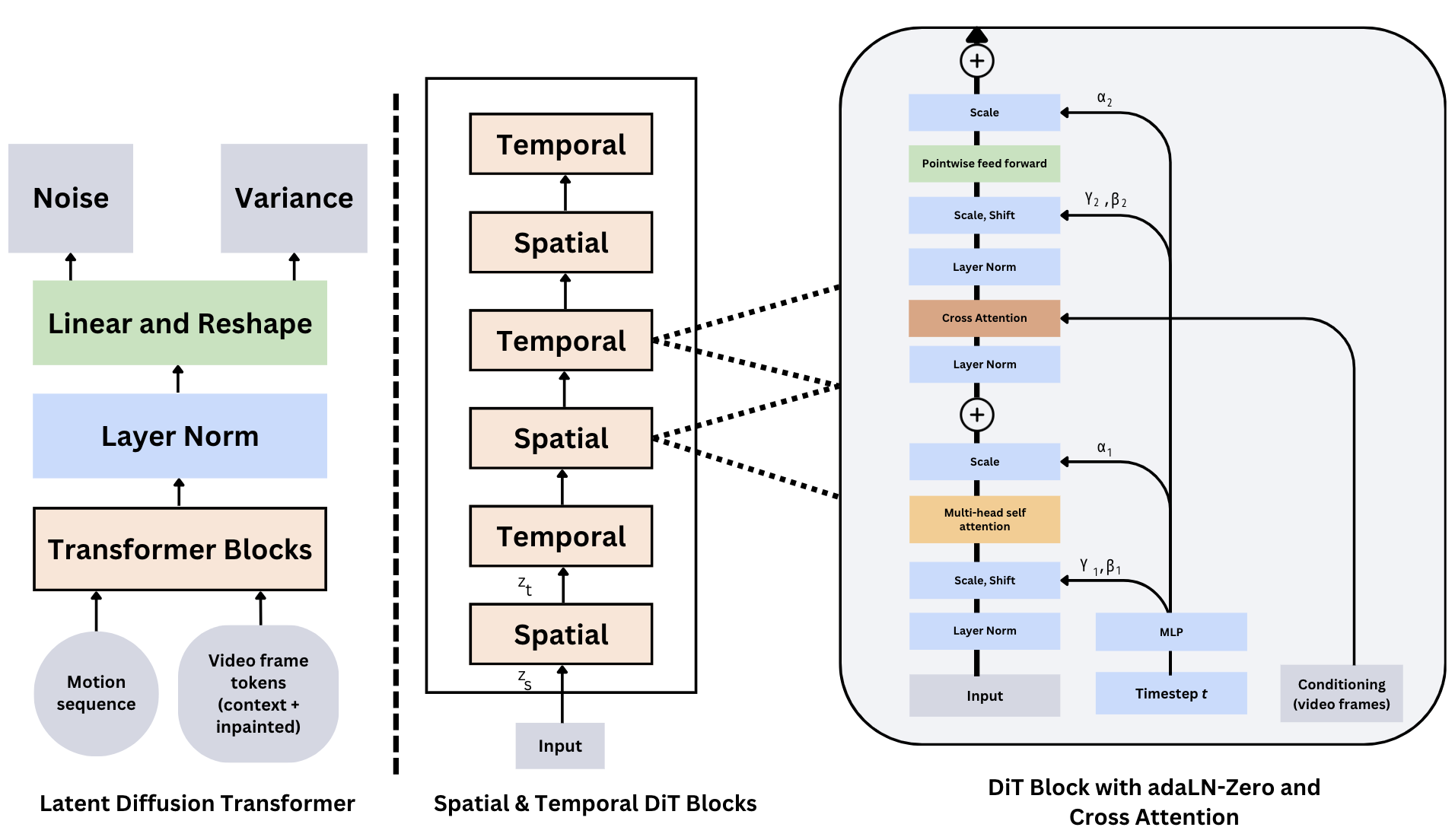}
  \caption{\textbf{The modified Diffusion Transformer (DiT) architecture.} Embedded video frames are taken in as conditioning via cross attention while the input motion latent vectors are fed through spatio-temporal DiT blocks.}
\end{figure}

\paragraph{Diffusion Model.}

Our core model is a modified DiT, aka diffusion transformer, that employs both zero-initialized adaptive layer normalization (adaLN-Zero) [\hyperlink{ref50}{50}, \hyperlink{ref53}{53}] and cross attention. It takes the full sequence of patch tokens from the cleaned video frames, with context frames, as conditioning, and uses in-context learning [\hyperlink{ref57}{57}, \hyperlink{ref58}{58}] to generate complete sequences of human motion vectors. When visualized, these vectors depict an AI avatar climbing the specified route with proper technique. The model incorporates efficient attention mechanisms, notably FlashAttention [\hyperlink{ref55}{55}, \hyperlink{ref56}{56}], to handle long video contexts effectively. Our transformer backbone comprises two distinct types of blocks: spatial transformer blocks and temporal transformer blocks. The spatial blocks capture information among tokens at the same temporal index, while the temporal blocks capture information across different time points in an interleaved fusion manner.

\section{Training Setup}

This section describes the training setup for our model.

\subsection{Training}

We initialize the final linear layer of our video conditional latent DiT model with zeros and otherwise use standard weight initialization techniques from ViT. We train our model with AdamW [\hyperlink{ref80}{80}, \hyperlink{ref81}{81}]. We use a constant learning rate of \(1 \times 10^{-4}\), no weight decay and a global batch size of 16 for 1500 epochs. We do no data augmentation. We did not find learning rate warmup nor regularization necessary to train our DiT, even for our new use case, to good performance. Even without these techniques, training was stable across our model config and we did not observe any severe loss spikes commonly seen when training transformers. Following common practice in the generative modeling literature, we maintain an exponential moving average (EMA) of DiT weights over training with a decay of 0.9999. All results reported use the EMA model. Our training hyperparameters are almost entirely retained from ADM [\hyperlink{ref78}{78}]. We did not tune learning rates, decay/warm-up schedules, Adam \(\beta_1/\beta_2\) or weight decays.

\subsection{Diffusion}
Diffusion is modeled as a Markov noising process, \(\{x_{1:N}^t\}_{t=0}^T\), where \((x_{1:N}^0)\) is drawn from the data distribution and
\[
q(x_{1:N}^t \mid x_{1:N}^{t-1}) = \mathcal{N}(\sqrt{\alpha_t} x_{1:N}^{t-1}, (1 - \alpha_t)I),
\]
where \(\alpha_t \in (0, 1)\) are constant hyperparameters. When \(\alpha_t\) is small enough, we can approximate \(x_{1:N}^T \sim \mathcal{N}(0, I)\). From here on, we use \(x^T\) to denote the full sequence at noising step \(t\).

Instead of predicting \(t\) [\hyperlink{ref7}{7}], we predict the signal itself [\hyperlink{ref79}{79}], i.e., \((\hat{x}^0 = G(x^t, t, c))\), with the simple objective [\hyperlink{ref7}{7}],
\[
L_{\text{simple}} = \mathbb{E}_{x_0 \sim q(x_0 \mid c), t \sim [1, T]} \left[\| x_0 - G(x^t, t, c) \|_2^2 \right].
\]

Furthermore, since our model performs cross-attention on patch tokens from DINOv2, it operates to an extent in DINOv2's Z-space. We retain diffusion hyperparameters from ADM [\hyperlink{ref78}{78}]; specifically, we use a \(t_{\text{max}} = 1000\) linear variance schedule ranging from \(1 \times 10^{-4}\) to \(2 \times 10^{-2}\), ADM’s parameterization of the covariance \(\Sigma_\theta\), and their method for embedding input timesteps.

\subsection{Loss}
We train our model using a mean squared error (MSE) loss between the sequences of ground truth vectors and predicted vectors. Each vector consists of the parameters \((\Theta_t = [\theta_t, \beta_t, \pi_t, b_t])\), where \(\theta_t\) and \(\beta_t\) are the SMPL pose and shape parameters, \(\pi_t\) is the camera translation, and \((b_t = [x_t, y_t, w_t, h_t])\) represents the bounding box.

\subsection{Compute}
We implement our \(\sim\)600M parameter diffusion transformer model in PyTorch and train it in mixed precision with PyTorch Distributed Data Parallel (DDP) on a single node of NVIDIA 8xA100 SXM (80GB) GPUs with a global batch size of 16.

\section{Evaluation}
We find that our architecture lays promising groundwork for world navigator models, which climbing walls provide an effective testbed for as they encompass various real-world challenges such as twists, turns, and complex biomechanical interactions. As expected, SABR-CLIMB generalizes well on climbing videos with excellent lighting, clean environment frames, and clear segmentation of the route with no awkward videography. With sufficient data and compute, the model effectively determines the scale and movement of the virtual avatar and identifies the appropriate climbing route. However, from a user experience perspective, the model currently provides general beta suggestions only for videos with optimal settings. It struggles with arbitrary video settings, which limits its consumer use case for now. We evaluate our model on a test dataset of 10 videos, each selected based on optimal input conditions for generalization. We upsample all output videos to 1440x2504 resolution for detailed qualitative analysis. Quantitatively, we measure route trajectory adherence—how well the AI avatar follows the correct route—and movement adherence—how accurately the AI avatar can climb the specified route.

\subsection{Evaluation Protocol}
We compare the avatar's full movement predictions with the user's original motion, using empty environment frames as input and 250 DDPM sampling steps. For videos showcasing different climbing methods on the same route, we feed in the first video and sample the diffusion model for a set period of time to check if the proposed motion includes movements present in the other videos. To visualize, we use the mean values of the virtual avatar's shape instead of the predicted ones. We ensure that the input videos are of high clarity and stable videography, avoiding random camera swings, to provide a clean environment for the model's predictions. 

To quantitatively evaluate the model's performance, we calculate two types of errors: movement adherence and route trajectory adherence. Movement adherence is measured using the mean square error loss between the ground truth motion tracks and the model's predicted motion. Route trajectory adherence is evaluated by counting the number of holds in the correct route and how many of these holds the AI avatar touches.

\subsection{Qualitative Evaluation}
We present qualitative examples around spatial, temporal, and depth understanding between the AI avatar and the climbing route video frames below.
\begin{figure}[ht]
  \centering
  \includegraphics[height=3cm]{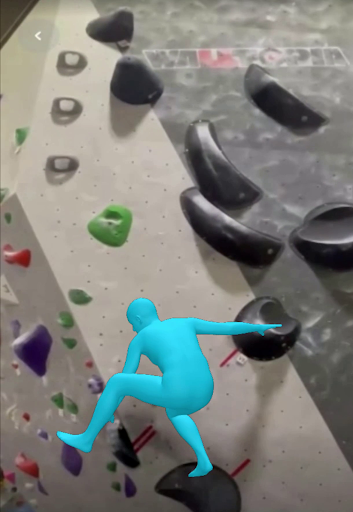}
  \includegraphics[height=3cm]{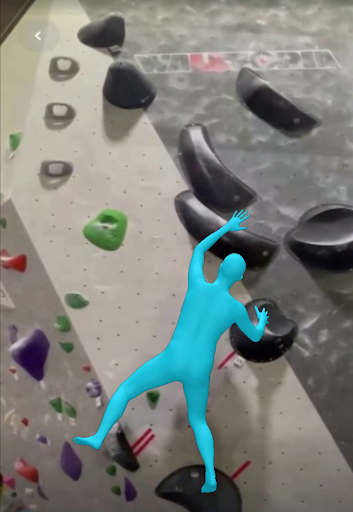}
  \includegraphics[height=3cm]{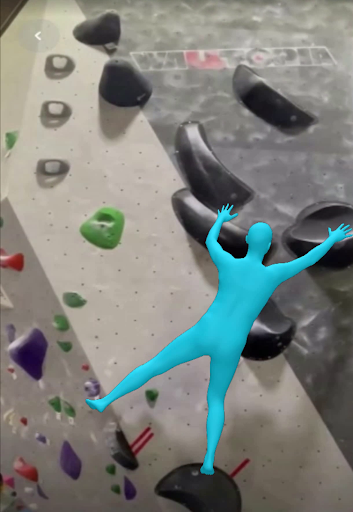}
  \includegraphics[height=3cm]{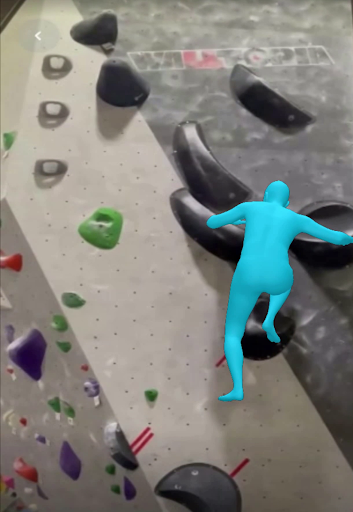}
  \includegraphics[height=3cm]{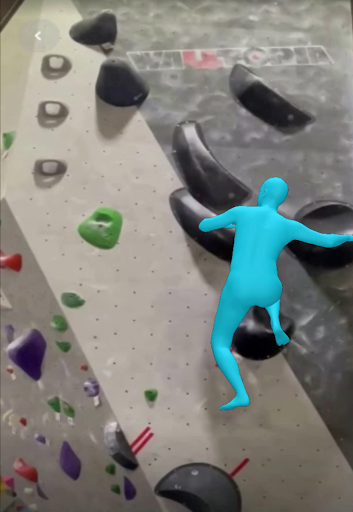}
  \includegraphics[height=3cm]{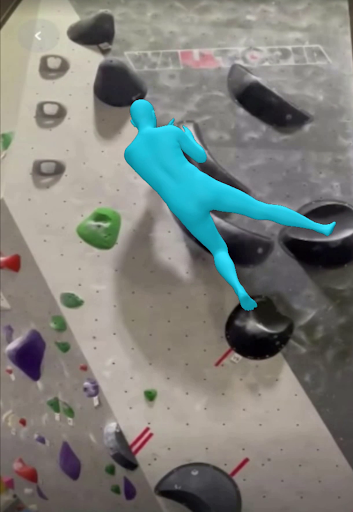}
  \caption{The AI avatar re-positions itself across two "slopers" on a black bouldering route.}
\end{figure}

\paragraph{Spatial Understanding.}
In Figure 6, we present qualitative results of SABR-CLIMB on a clear input video of a black bouldering route. These results indicate SABR-CLIMB's promising capability to understand the relationship between various holds and human motion.

\begin{figure}[ht]
  \centering
  \includegraphics[height=1.95cm]{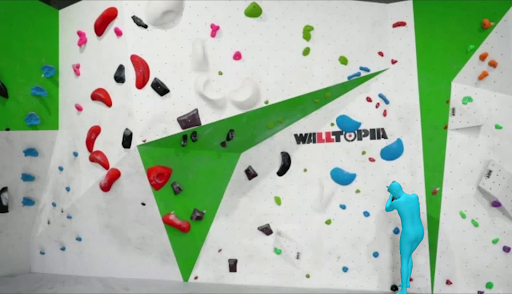}
  \includegraphics[height=1.95cm]{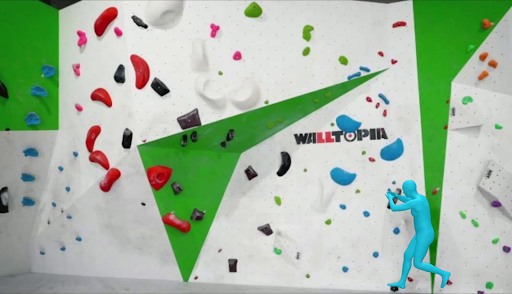}
  \includegraphics[height=1.95cm]{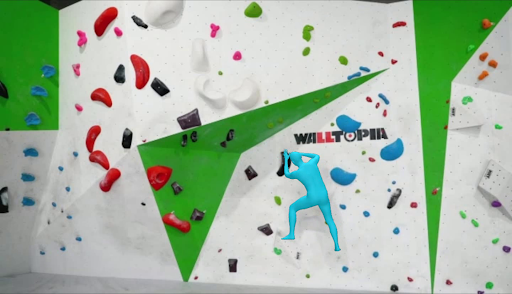}
  \includegraphics[height=1.95cm]{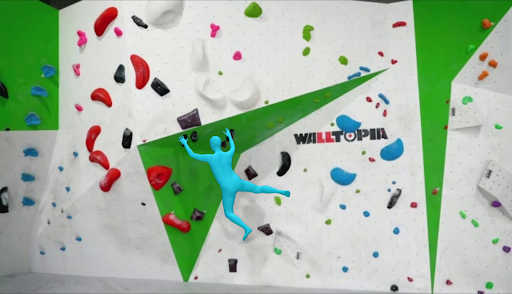}
  \caption{The AI avatar does a "dyno" on a black bouldering route.}
\end{figure}

\paragraph{Temporal Understanding.}
In Figure 7, we provide an example of SABR-CLIMB's understanding of human motion, specifically the velocity of limb movements. This demonstrates the model's ability to generate realistic motion sequences over time. Since the diffusion process is on entire motion sequences with conditioning on entire videos, SABR-CLIMB can learn the correlations between different types of movements. For example, it understands that performing a "dyno" (aka dynamic move) in the above setting involves generating momentum with the legs at the start.

\begin{figure}[ht]
  \centering
  \includegraphics[height=4cm]{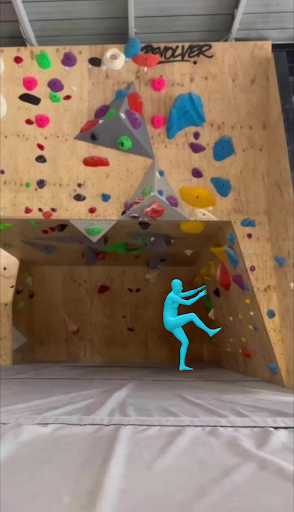}
  \includegraphics[height=4cm]{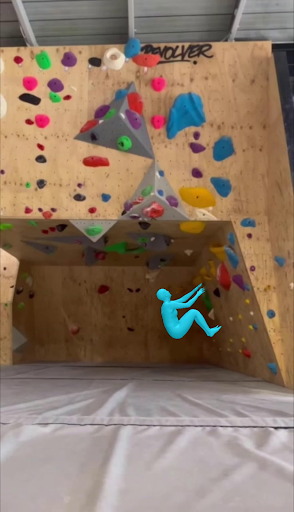}
  \includegraphics[height=4cm]{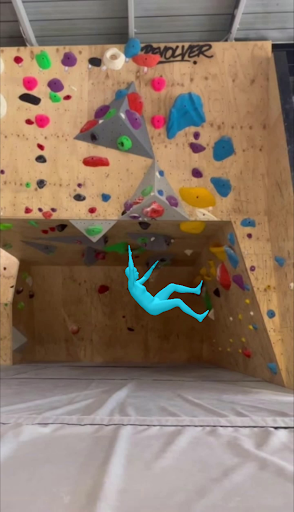}
  \includegraphics[height=4cm]{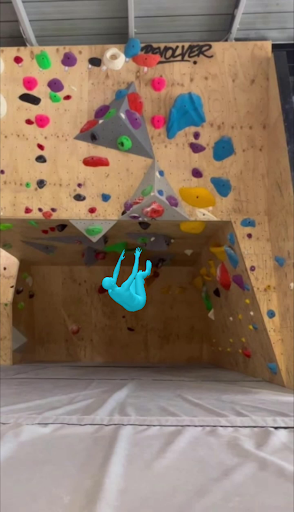}
  \includegraphics[height=4cm]{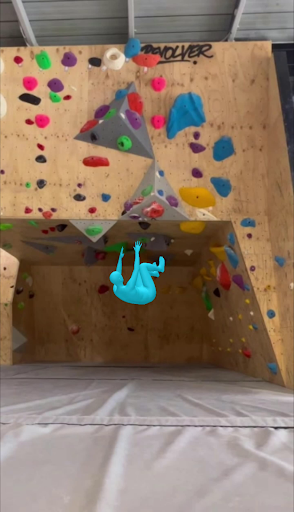}
  \caption{The AI avatar climbs a green route inside of a bouldering cave.}
\end{figure}

\paragraph{Depth Understanding.}
In Figure 8, we showcase SABR-CLIMB's understanding of depth in a cave-like setting. The model accurately scales the virtual avatars to match the environment's depth, indicating effective depth perception.

\begin{figure}[ht]
  \centering
  \includegraphics[height=4cm]{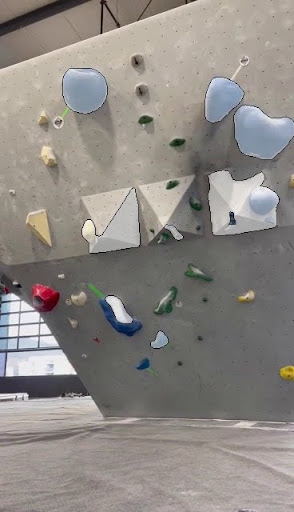}
  \includegraphics[height=4cm]{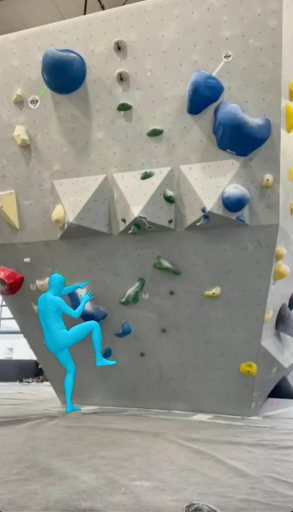}
  \includegraphics[height=4cm]{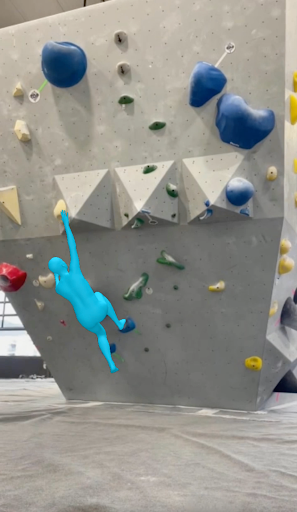}
  \includegraphics[height=4cm]{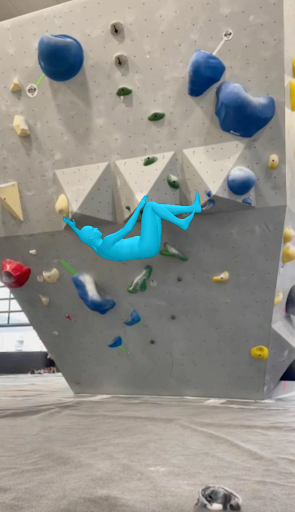}
  \includegraphics[height=4cm]{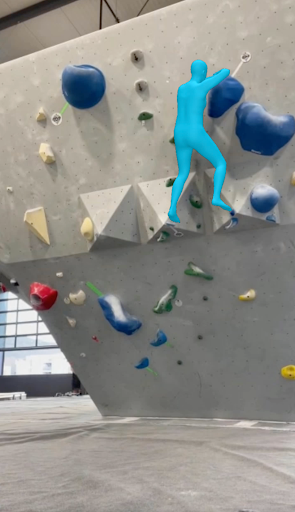}
  \caption{The AI avatar climbs a blue route on an indoor bouldering wall.}
\end{figure}

\paragraph{Context Understanding.}
In Figure 9, we show that strict annotation of the route provides enough contextual information for SABR-CLIMB to determine the correct climbing route.

\subsection{Quantitative Evaluation}
\begin{figure}[ht]
  \centering
  \includegraphics[height=5cm]{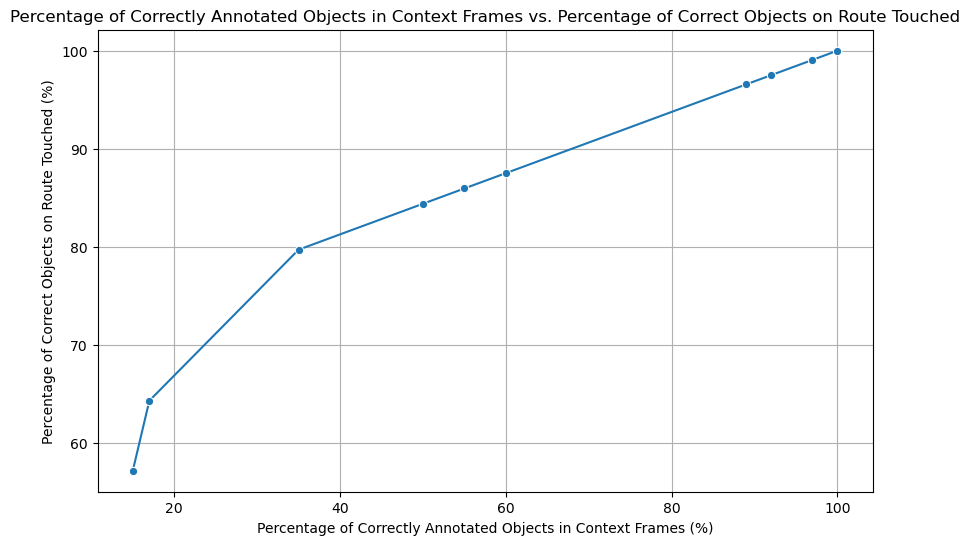}
  \caption{\textbf{Trajectory adherence based on number of correctly annotated objects in the context frames.} We show that the number of correctly annotated objects is a significant driver for accurate route following.}
\end{figure}

\paragraph{Trajectory Adherence.}
Figure 10 presents a visual comparison of the trajectory adherence of our avatar generation model based on the number of annotated objects in the correct route within the context frames. Our experiments reveal that the primary factor influencing trajectory adherence is the quantity of annotated objects rather than the strict number of context video frames. Some videos require more context frames to adequately cover the full route due to movement within the frames. Our findings indicate that even with only 20\% of the context frames annotated, our model demonstrates substantial route-following capability, making it suitable for motion completion. For instance, when a user inputs a failed video, the avatar can accurately determine the correct route.

\begin{figure}[ht]
  \centering
  \includegraphics[height=5cm]{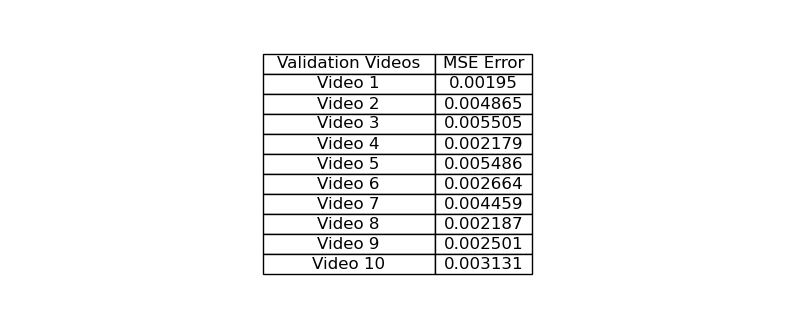}
  \caption{\textbf{Movement adherence based on mean square error loss between predictions and ground truth motion.} We demonstrate that our model achieves a low validation loss. Given that the predictions encompass full video sequences, including pose, body shape, camera, and bounding boxes, this low loss indicates highly accurate movements throughout the entire video.}
\end{figure}

\paragraph{Movement Adherence.}

Figure 11 presents a table showing the mean square error loss between the predicted video motion vectors and the ground truth. We find that a lower overall loss directly correlates with better movement adherence. Our motion vectors, which primarily consist of pose, body shape, and camera parameters, together form the body mesh model, indicating that accurate predictions in these areas lead to improved overall movement fidelity.

\subsection{Scaling Studies}
\begin{figure}[ht]
  \centering
  \includegraphics[height=4cm]{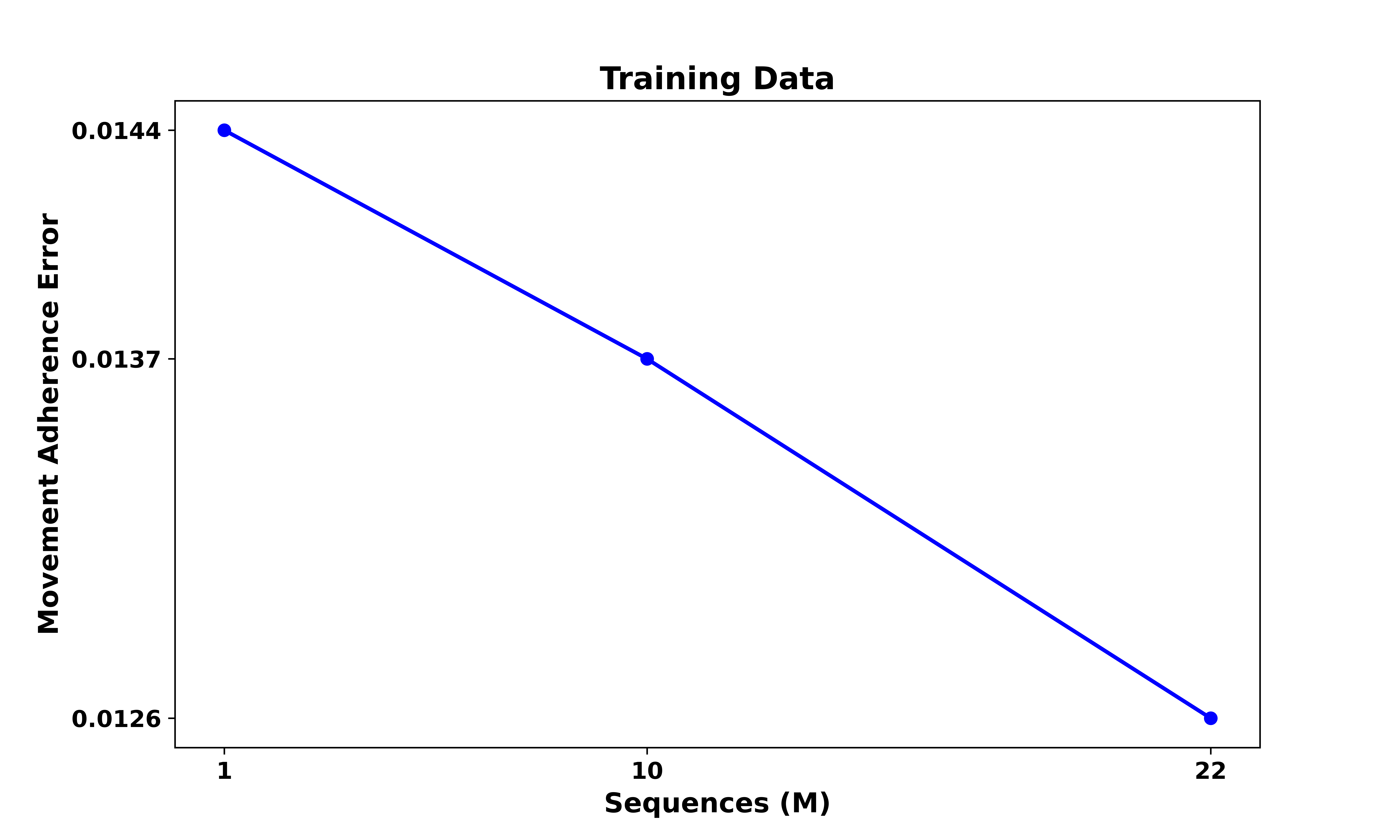}
  \includegraphics[height=4cm]{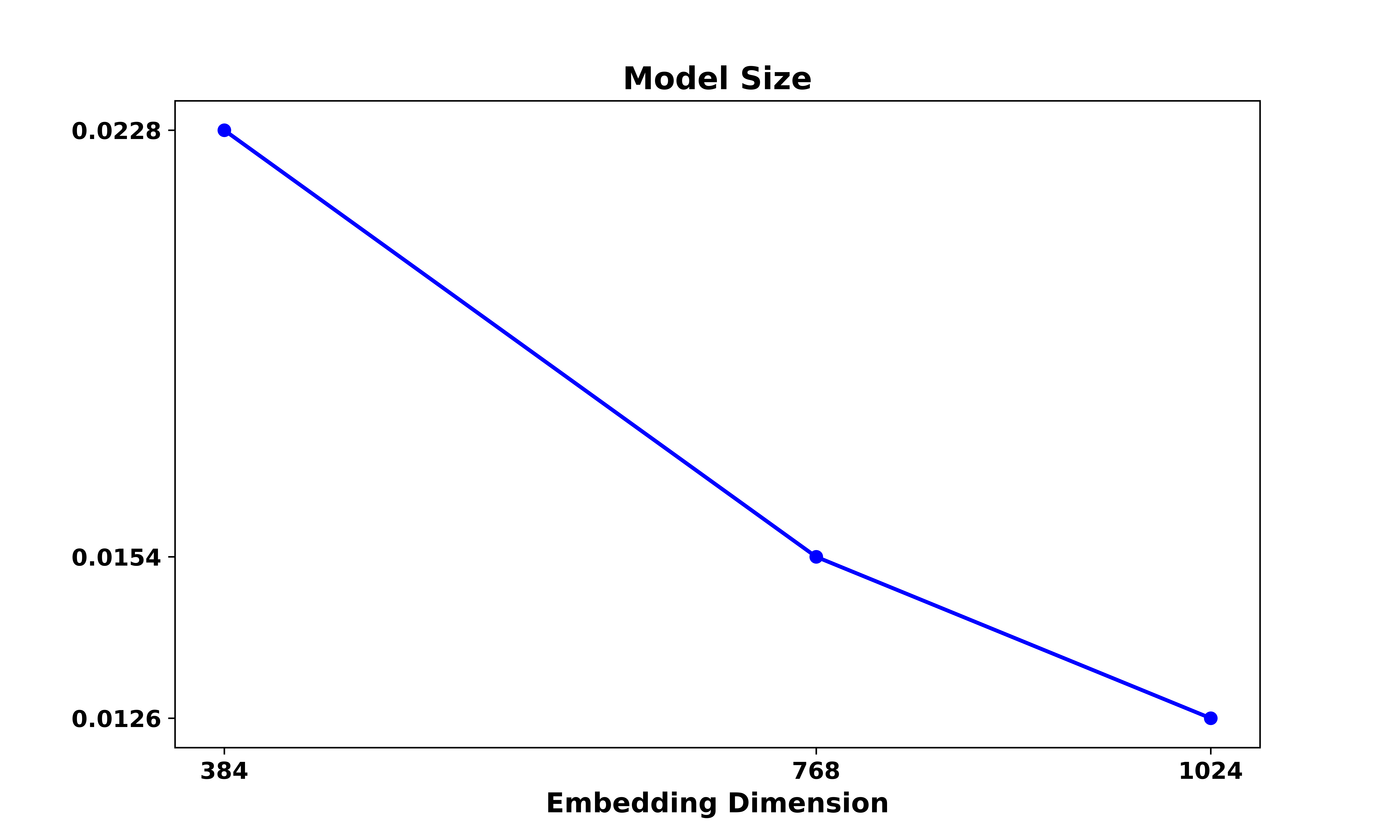}
  \caption{\textbf{Scaling studies.} We find that our approach scales with the number of environment sequences in the training dataset (left) and larger models (right).}
\end{figure}

\paragraph{Training Data.} We study the scaling of our model's performance for 10 epochs by increasing the size of the training dataset for our base 600M parameter model. We find that training on more environment sequences reduces movement adherence error, which is a positive signal for increased performance when training on larger datasets.

\paragraph{Model Size.} We compare models with varying numbers of parameters for 10 epochs on our full NAV-22M dataset by adjusting the embedding dimension (384, 768, 1024), the number of attention heads (6, 12, 16), and the number of transformer blocks (12, 12, 24) respectively. We observe that movement adherence error decreases monotonically with increasing model size. For the 1024 embedding size, we use the large version of our video frame encoder DINOv2, while for the 768 and 384 embedding dimensions, we use the medium and small versions, respectively. This variation affects model quality due to the differing quality of input patch tokens. Our final model configuration, featuring a 1024 embedding dimension, 16 attention heads, and 24 transformer blocks, demonstrates the best performance.

\subsection{Limitations}
While our model shows promising results, there are still some limitations. The 3D pose tracker (PHALP) we utilized was not always perfect, resulting in inherently flawed ground truth data. Additionally, the video inpainting model occasionally produced artifacts that occluded key holds on the climbing wall. Training video models, especially the diffusion model that conditioned on all patch tokens from all frames, was challenging due to the computational intensity and slow preprocessing of entire videos compared to text or images, limiting our training duration. Collecting the dataset was similarly difficult, as running both the 3D pose tracker and the video inpainting model was computationally demanding. Consequently, our current model iteration struggles in imperfect environments and does not generalize well to unseen types of climbing walls. We provide qualitative examples of these limitations below. Future improvements could address these issues by enhancing the data processing pipeline to expedite data collection and improving video processing infrastructure to increase training iterations per second. Additionally, refining the model architecture by tokenizing motion to some extent, as discussed in the conclusion section, could further mitigate these challenges.

\begin{figure}[ht]
  \centering
  \includegraphics[height=2.5cm]{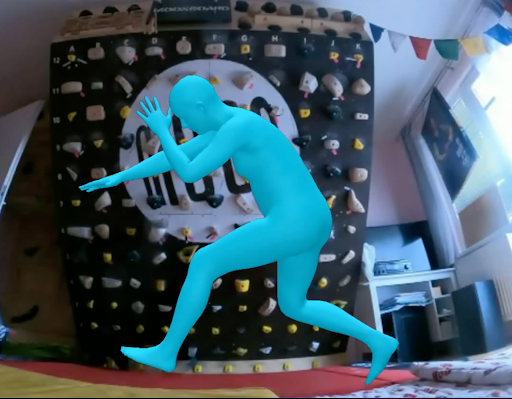}
  \includegraphics[height=2.5cm]{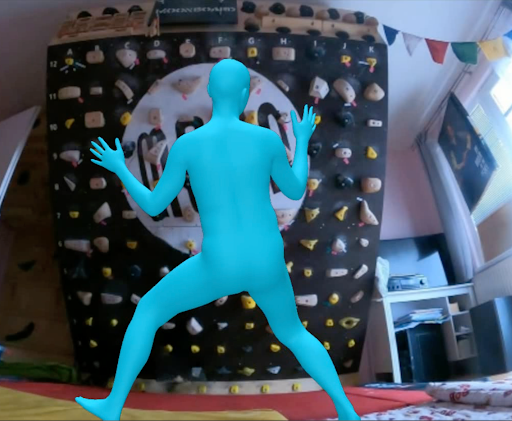}
  \includegraphics[height=2.5cm]{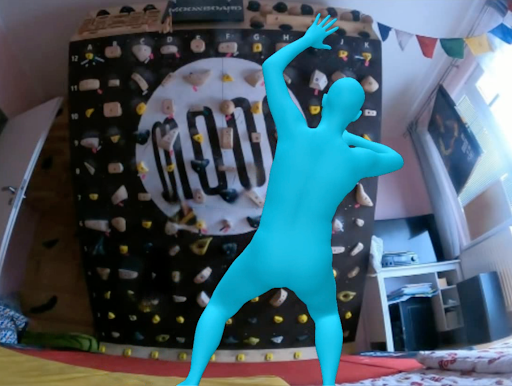}
  \includegraphics[height=2.5cm]{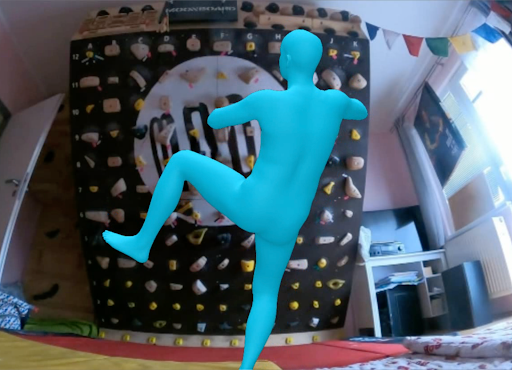}
  \caption{The AI avatar struggles to position itself across an unseen type of MoonBoard.}
\end{figure}
\paragraph{Out of distribution avatar shape and scale.}
SABR-CLIMB struggles with videos that fall outside its training distribution or videos with unnatural routes (Fig. 13), resulting in avatars with incorrect scales and shapes. This can significantly affect the accuracy of the predicted movements and overall usability in diverse climbing scenarios.

\begin{figure}[ht]
  \centering
  \includegraphics[height=4cm]{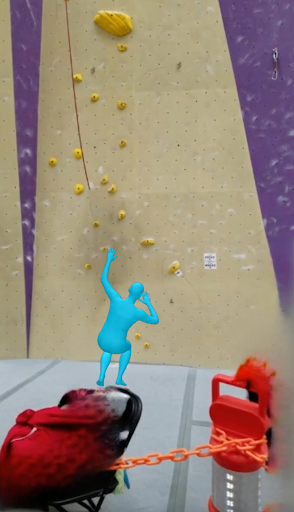}
  \includegraphics[height=4cm]{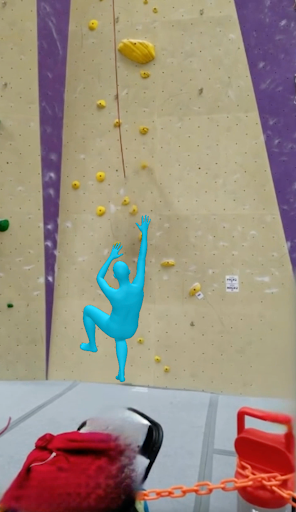}
  \includegraphics[height=4cm]{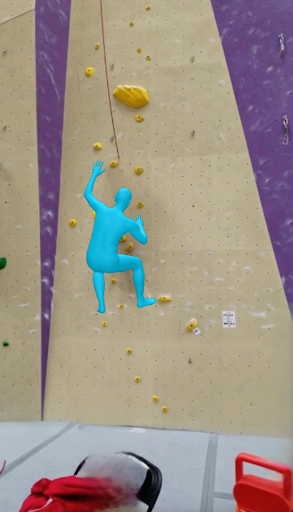}
  \includegraphics[height=4cm]{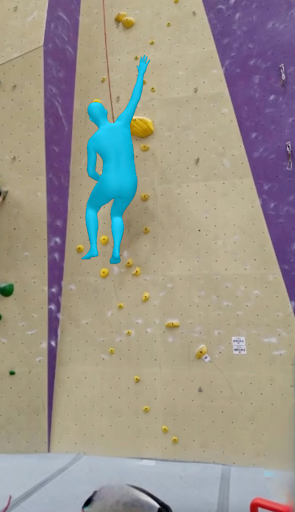}
  \includegraphics[height=4cm]{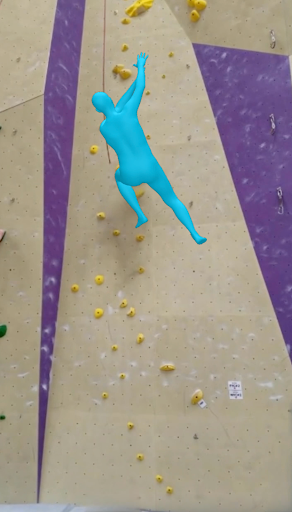}
  \caption{The AI avatar struggles to grip the proper holds and move with the proper speed on a speed climbing route.}
\end{figure}
\paragraph{Fine-grained spatial understanding.}
The model sometimes fails to accurately recognize and interact with small holds and volumes (Fig. 14). In cases where rocks are too small or in configurations that the model has not frequently encountered, SABR-CLIMB may miss them entirely or fail to generate appropriate movements, leading to unrealistic or unsafe motion predictions.

\begin{figure}[ht]
  \centering
  \includegraphics[height=1.8cm]{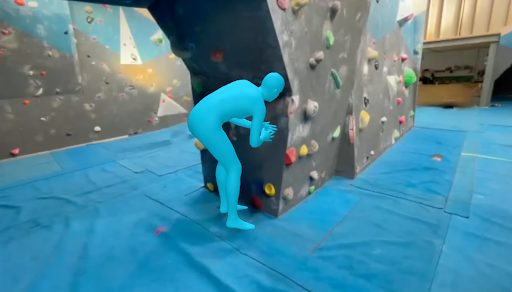}
  \includegraphics[height=1.8cm]{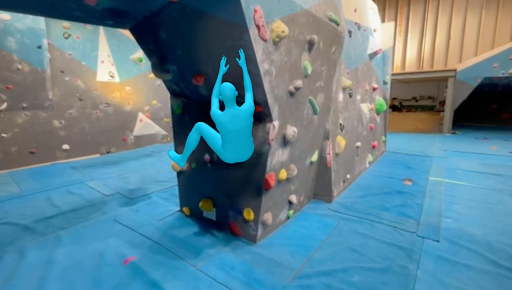}
  \includegraphics[height=1.8cm]{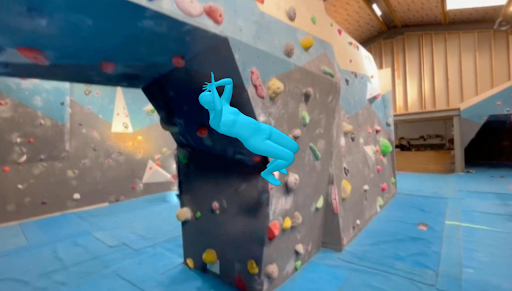}
  \includegraphics[height=1.8cm]{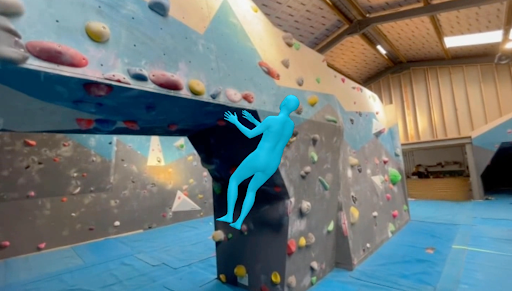}
  \caption{The AI avatar struggles to move in an arched route.}
\end{figure}
\paragraph{Arbitrary videography.}
The best outputs from SABR-CLIMB occur with input videos where the environment and context are crystal clear. Unfortunately, most in-the-wild videos are very noisy, with inconsistent videography that includes twists and turns (Fig. 15), making motion prediction challenging. Downsampling the resolution to handle these inconsistencies negatively affects spatial capabilities in some videos.

\paragraph{Video duration.}
SABR-CLIMB can generate simulations up to 45 seconds. However, some climbs may exceed this duration, causing the model to lose important context needed for predicting accurate future movements. This limitation restricts the model's applicability to longer or more complex climbing routes.

\subsection{Impact of Architecture}
Initially, we tried using a transformer decoder with geometric losses. However, even small changes in pose parameters resulted in unnatural movements, indicating difficulties in tokenizing human motion. Additionally, the loss landscape was fraught with local minima, complicating the training process. We also conducted experiments with classifier-free guidance but observed no tangible improvement in the model outputs. We did not experiment with different backbones as we wanted to optimize for efficiency and scalability, which the transformer is best for. Ultimately, diffusion models outperformed other architectures due to their ability to effectively model the input data distribution and handle the many-to-many problem inherent in human motion prediction.

\subsection{Impact of Video Encoder}
We compared DINOv2 and CLIP \hyperlink{ref36}{[36]}, finding that DINOv2 provided more consistent embeddings. The large model of DINOv2 was best-suited for our use case because it captured the fine-grained details of holds and wall types the most effectively while still being computationally efficient enough for our model to process with the compute we had.

\subsection{Impact of Video Resolution and FPS}
Higher resolution yields better results but is much more computationally intensive. Our environment model is an encoder-decoder transformer model with a specific stride, requiring the width and height of each video frame to be divisible by 4. Most rock climbing videos have a 16:9 aspect ratio, so we aimed for a downsampled resolution that fits the stride of our inpainting model, maintains computational efficiency, and preserves the original aspect ratio as closely as possible. We experimented with a resolution of 480x864, but found that the video processing pipeline remained too slow, taking multiple hours per video.We opted for 368x640 for efficiency. For FPS, we tested 30, 24, 20, and 10. We chose 24 FPS as it balanced fast 3D pose tracking and maintainable motion quality. Higher FPS improved smoothness but was computationally intensive, while lower FPS resulted in choppy motion.

\subsection{Impact of Video Preprocessing Pipeline}
For 3D pose tracking, we used PHALP, the state of the art. We tested various other video inpainting methods [\hyperlink{ref26}{26}, \hyperlink{ref27}{27}, \hyperlink{ref28}{28}] but found that FGT performed best due to its minimal artifacts in slow-moving climbing videos. Other models introduced significant artifacts. We modified FGT to handle arbitrary video lengths without artifacts, enhancing its suitability for our dataset and use case.

\section{Conclusion}
In this work, we have presented a deep neural architecture that learns human biomechanics to navigate 2D RGB videos in the rock climbing domain. Even with limited data, our model exhibits instances of spatial, temporal, and motion reasoning - critical for applications in world navigation. As with any generative modeling method, the quality of the model is directly tied to the scale and quality of the training data and the duration of the training. Current SMPL body models are fairly constraining, focusing primarily on shape and size of the body only. Our goal is to simulate entire human models by including granular movements of the hand \hyperlink{ref60}{[60]} and feet as well as capturing a broader range of human biomechanics and even broader human behavior. We are also exploring alternative methods for motion control [\hyperlink{ref62}{62}, \hyperlink{ref63}{63}] and advancements in post-training processes, different pipelines, and end-to-end fusion models \hyperlink{ref35}{[35]}. Improving vector quantized variational autoencoders, VQ-VAEs \hyperlink{ref2}{[2]} and other forms of encoders, will enable better scene and motion tokenization [\hyperlink{ref64}{64}, \hyperlink{ref65}{65}, \hyperlink{ref66}{66}], allowing us to purely use autoregressive models such as transformers, which have seen significant research advancements in efficiency and scalability in adjacent domains \hyperlink{ref61}{[61]}. While diffusion models can model input data distributions better than other models, they are challenging and expensive to train and can be unreliable. By enhancing VQ-VAE capabilities, we can leverage the extensive research on autoregressive models to improve efficiency and reliability. Overall, future work will focus on scaling up data and computational resources, refining the model architecture, and exploring new methods to enhance the realism and applicability of the virtual avatar generation model. 

Our work here is a plausible direction for general purpose robotics because robots can use virtual avatars to learn directly from 2D input videos and extrapolate sensorimotor trajectories in 3D. By converting reference kinematic motion in 2D videos from a virtual avatar generation model into physically executable actions, robots can acquire complex skills and movements beyond simple tasks like cooking or picking up objects. This capability extends to more intricate settings such as playing sports and collaborating with humans in rehabilitation therapy and healthcare.

\section*{References}

{
\small

\hypertarget{ref1}{[1]} Tevet, Guy, Raab, Sigal, Gordon, Brian, Shafir, Yoni, Cohen-Or, Daniel, \& Bermano, Amit Haim. "Human Motion Diffusion Model." The Eleventh International Conference on Learning Representations, 2023. https://openreview.net/forum?id=SJ1kSyO2jwu.

\hypertarget{ref2}{[2]} Diederik P Kingma \& Max Welling. Auto-encoding variational bayes. arXiv preprint arXiv:1312.6114, 2013.

\hypertarget{ref3}{[3]} Chuan Guo, Xinxin Zuo, Sen Wang, Shihao Zou, Qingyao Sun, Annan Deng, Minglun Gong, \& Li Cheng. Action2motion: Conditioned generation of 3d human motions. In Proceedings of the 28th ACM International Conference on Multimedia, pp. 2021–2029, 2020.

\hypertarget{ref4}{[4]} Chuan Guo, Shihao Zou, Xinxin Zuo, Sen Wang, Wei Ji, Xingyu Li, \& Li Cheng. Generating diverse and natural 3d human motions from text. In Proceedings of the IEEE/CVF Conference on Computer Vision and Pattern Recognition, pp. 5152–5161, 2022a.

\hypertarget{ref5}{[5]} Alejandro Hernandez, Jurgen Gall, \& Francesc Moreno-Noguer. Human motion prediction via spatio-temporal inpainting. In Proceedings of the IEEE/CVF International Conference on Computer Vision, pp. 7134–7143, 2019.

\hypertarget{ref6}{[6]} Jonathan Ho \& Tim Salimans. Classifier-free diffusion guidance. arXiv preprint arXiv:2207.12598, 2022.

\hypertarget{ref7}{[7]} Jonathan Ho, Ajay Jain,\& Pieter Abbeel. Denoising diffusion probabilistic models. Advances in Neural Information Processing Systems, 33:6840–6851, 2020.

\hypertarget{ref8}{[8]} Jonathan Ho, Tim Salimans, Alexey Gritsenko, William Chan, Mohammad Norouzi, \& David J Fleet. Video diffusion models. arXiv preprint arXiv:2204.03458, 2022.

\hypertarget{ref9}{[9]} Mathis Petrovich, Michael J. Black, \& Gul Varol. TEMOS: Generating diverse human motions from textual descriptions. In European Conference on Computer Vision (ECCV), 2022.

\hypertarget{ref10}{[10]} Chaitanya Ahuja \& Louis-Philippe Morency. Language2pose: Natural language grounded pose forecasting. In 2019 International Conference on 3D Vision (3DV), pp. 719–728. IEEE, 2019.

\hypertarget{ref11}{[11]} Emre Aksan, Manuel Kaufmann, Peng Cao, \& Otmar Hilliges. A spatio-temporal transformer for 3d human motion prediction. In 2021 International Conference on 3D Vision (3DV), pp. 565–574. IEEE, 2021.

\hypertarget{ref12}{[12]} Uttaran Bhattacharya, Nicholas Rewkowski, Abhishek Banerjee, Pooja Guhan, Aniket Bera, \& Dinesh Manocha. Text2gestures: A transformer-based network for generating emotive body gestures for virtual agents. In 2021 IEEE Virtual Reality and 3D User Interfaces (VR), pp. 1–10. IEEE, 2021.

\hypertarget{ref13}{[13]} Prafulla Dhariwal \& Alexander Nichol. Diffusion models beat gans on image synthesis. Advances in Neural Information Processing Systems, 34:8780–8794, 2021.

\hypertarget{ref14}{[14]} Katerina Fragkiadaki, Sergey Levine, Panna Felsen, \& Jitendra Malik. Recurrent network models for human dynamics. In Proceedings of the IEEE international conference on computer vision, pp. 4346–4354, 2015.

\hypertarget{ref15}{[15]} Jonathan Ho, Ajay Jain, \& Pieter Abbeel. Denoising diffusion probabilistic models. Advances in Neural Information Processing Systems, 33:6840–6851, 2020.

\hypertarget{ref16}{[16]} Alex Nichol, Prafulla Dhariwal, Aditya Ramesh, Pranav Shyam, Pamela Mishkin, Bob McGrew, Ilya Sutskever, \& Mark Chen. Glide: Towards photorealistic image generation and editing with text-guided diffusion models. arXiv preprint arXiv:2112.10741, 2021.

\hypertarget{ref17}{[17]} Olaf Ronneberger, Philipp Fischer, \& Thomas Brox. U-net: Convolutional networks for biomedical image segmentation. In International Conference on Medical image computing and computerassisted intervention, pp. 234–241. Springer, 2015.

\hypertarget{ref18}{[18]} Yang Song, Jascha Sohl-Dickstein, Diederik P Kingma, Abhishek Kumar, Stefano Ermon, \& Ben Poole. Score-based generative modeling through stochastic differential equations. arXiv preprint arXiv:2011.13456, 2020b.

\hypertarget{ref19}{[19]} Ashish Vaswani, Noam Shazeer, Niki Parmar, Jakob Uszkoreit, Llion Jones, Aidan N Gomez, Łukasz Kaiser, \& Illia Polosukhin. Attention is all you need. Advances in neural information processing systems, 30, 2017.

\hypertarget{ref20}{[20]} Jathushan Rajasegaran, Georgios Pavlakos, Angjoo Kanazawa, \& Jitendra Malik. Tracking people by predicting 3D appearance, location and pose. In CVPR, 2022.

\hypertarget{ref21}{[21]} Matthew Loper, Naureen Mahmood, Javier Romero, Gerard Pons-Moll, \& Michael J Black. SMPL: A skinned multiperson linear model. ACM transactions on graphics (TOG), 34(6):1–16, 2015.

\hypertarget{ref22}{[22]} Goel, Shubham and Pavlakos, Georgios and Rajasegaran, Jathushan and Kanazawa, Angjoo \& Malik, Jitendra. Humans in 4D: Reconstructing and Tracking Humans with Transformers. In ICCV, 2023.

\hypertarget{ref23}{[23]} Darcet, Timothée, Oquab, Maxime, Mairal, Julien, \& Bojanowski, Piotr. "Vision Transformers Need Registers." arXiv:2309.16588, 2023.

\hypertarget{ref24}{[24]} Oquab, Maxime, Darcet, Timothée, Moutakanni, Theo, Vo, Huy V., Szafraniec, Marc, Khalidov, Vasil, Fernandez, Pierre, Haziza, Daniel, Massa, Francisco, El-Nouby, Alaaeldin, Howes, Russell, Huang, Po-Yao, Xu, Hu, Sharma, Vasu, Li, Shang-Wen, Galuba, Wojciech, Rabbat, Mike, Assran, Mido, Ballas, Nicolas, Synnaeve, Gabriel, Misra, Ishan, Jegou, Herve, Mairal, Julien, Labatut, Patrick, Joulin, Armand, \& Bojanowski, Piotr. "DINOv2: Learning Robust Visual Features without Supervision." arXiv:2304.07193, 2023.

\hypertarget{ref25}{[25]} Zhang, Kaidong, Fu, Jingjing, \& Liu, Dong. "Flow-Guided Transformer for Video Inpainting." European Conference on Computer Vision, Springer, 2022, pp. 74-90

\hypertarget{ref26}{[26]} Zhang, Kaidong, Fu, Jingjing, \& Liu, Dong. "Inertia-Guided Flow Completion and Style Fusion for Video Inpainting." Proceedings of the IEEE/CVF Conference on Computer Vision and Pattern Recognition (CVPR), June 2022, pp. 5982-5991.

\hypertarget{ref27}{[27]} Li, Zhen, Lu, Cheng-Ze, Qin, Jianhua, Guo, Chun-Le, \& Cheng, Ming-Ming. "Towards An End-to-End Framework for Flow-Guided Video Inpainting." IEEE Conference on Computer Vision and Pattern Recognition (CVPR), 2022.

\hypertarget{ref28}{[28]} Zhou, Shangchen, Li, Chongyi, Chan, Kelvin C.K., \& Loy, Chen Change. "ProPainter: Improving Propagation and Transformer for Video Inpainting." Proceedings of IEEE International Conference on Computer Vision (ICCV), 2023.

\hypertarget{ref29}{[29]} Kirillov, Alexander, Mintun, Eric, Ravi, Nikhila, Mao, Hanzi, Rolland, Chloe, Gustafson, Laura, Xiao, Tete, Whitehead, Spencer, Berg, Alexander C., Lo, Wan-Yen, Dollár, Piotr, \& Girshick, Ross. "Segment Anything." arXiv:2304.02643, 2023.

\hypertarget{ref30}{[30]} Cheng, Yangming, Li, Liulei, Xu, Yuanyou, Li, Xiaodi, Yang, Zongxin, Wang, Wenguan, \& Yang, Yi. "Segment and Track Anything." arXiv preprint arXiv:2305.06558, 2023.

\hypertarget{ref31}{[31]} Yang, Zongxin, \& Yang, Yi. "Decoupling Features in Hierarchical Propagation for Video Object Segmentation." Advances in Neural Information Processing Systems (NeurIPS), 2022.

\hypertarget{ref32}{[32]} Yang, Zongxin, Wei, Yunchao, \& Yang, Yi. "Associating Objects with Transformers for Video Object Segmentation." Advances in Neural Information Processing Systems (NeurIPS), 2021.

\hypertarget{ref33}{[33]} Yang, Zhendong, Zeng, Ailing, Yuan, Chun, \& Li, Yu. "Effective Whole-Body Pose Estimation with Two-Stages Distillation." Proceedings of the IEEE/CVF International Conference on Computer Vision, 2023, pp. 4210-4220.

\hypertarget{ref34}{[34]} van den Oord, Aäron, Vinyals, Oriol, \& Kavukcuoglu, Koray. "Neural Discrete Representation Learning." arXiv:1711.00937, 2018.

\hypertarget{ref35}{[35]} Chameleon Team. "Chameleon: Mixed-Modal Early-Fusion Foundation Models." arXiv:2405.09818, 2024. 

\hypertarget{ref36}{[36]} Radford, Alec, Kim, Jong Wook, Hallacy, Chris, Ramesh, Aditya, Goh, Gabriel, Agarwal, Sandhini, Sastry, Girish, Askell, Amanda, Mishkin, Pamela, Clark, Jack, Krueger, Gretchen, \& Sutskever, Ilya. "Learning Transferable Visual Models From Natural Language Supervision." arXiv:2103.00020, 2021.

\hypertarget{ref37}{[37]} Bar-Tal, Omer, Chefer, Hila, Tov, Omer, Herrmann, Charles, Paiss, Roni, Zada, Shiran, Ephrat, Ariel, Hur, Junhwa, Liu, Guanghui, Raj, Amit, Li, Yuanzhen, Rubinstein, Michael, Michaeli, Tomer, Wang, Oliver, Sun, Deqing, Dekel, Tali, \& Mosseri, Inbar. "Lumiere: A Space-Time Diffusion Model for Video Generation." arXiv:2401.12945, 2024.

\hypertarget{ref38}{[38]} Zhou, Zhi-Hua. "A Theoretical Perspective of Machine Learning with Computational Resource Concerns." arXiv:2305.02217, 2023.

\hypertarget{ref39}{[39]} Gupta, Agrim, Yu, Lijun, Sohn, Kihyuk, Gu, Xiuye, Hahn, Meera, Fei-Fei, Li, Essa, Irfan, Jiang, Lu, \& Lezama, José. "Photorealistic Video Generation with Diffusion Models." arXiv:2312.06662, 2023.

\hypertarget{ref40}{[40]} Felix G Harvey, Mike Yurick, Derek Nowrouzezahrai, \& Christopher Pal. Robust motion in- ´ betweening. ACM Transactions on Graphics (TOG), 39(4):60–1, 2020.

\hypertarget{ref41}{[41]} Wen Guo, Yuming Du, Xi Shen, Vincent Lepetit, Xavier Alameda-Pineda, \& Francesc MorenoNoguer. Back to mlp: A simple baseline for human motion prediction. arXiv preprint arXiv:2207.01567, 2022b.

\hypertarget{ref42}{[42]} Mingyuan Zhang, Zhongang Cai, Liang Pan, Fangzhou Hong, Xinying Guo, Lei Yang, \& Ziwei Liu. Motiondiffuse: Text-driven human motion generation with diffusion model. arXiv preprint arXiv:2208.15001, 2022.

\hypertarget{ref43}{[43]} Kyunghyun Cho, Bart Van Merrienboer, Caglar Gulcehre, Dzmitry Bahdanau, Fethi Bougares, Hol- ¨ ger Schwenk, \& Yoshua Bengio. Learning phrase representations using rnn encoder-decoder for statistical machine translation. arXiv preprint arXiv:1406.1078, 2014.

\hypertarget{ref44}{[44]} Daniel Holden, Jun Saito, \& Taku Komura. A deep learning framework for character motion synthesis and editing. ACM Transactions on Graphics (TOG), 35(4):1–11, 2016.

\hypertarget{ref45}{[45]} Pablo Cervantes, Yusuke Sekikawa, Ikuro Sato, \& Koichi Shinoda. Implicit neural representations for variable length human motion generation. arXiv preprint arXiv:2203.13694, 2022.

\hypertarget{ref46}{[46]} Ruilong Li, Shan Yang, David A. Ross, \& Angjoo Kanazawa. Ai choreographer: Music conditioned 3d dance generation with aist++. In The IEEE International Conference on Computer Vision (ICCV), 2021.

\hypertarget{ref47}{[47]} A Aristidou, A Yiannakidis, K Aberman, D Cohen-Or, A Shamir, \& Y Chrysanthou. Rhythm is a dancer: Music-driven motion synthesis with global structure. IEEE Transactions on Visualization and Computer Graphics, 2022.

\hypertarget{ref48}{[48]} Chitwan Saharia, William Chan, Huiwen Chang, Chris Lee, Jonathan Ho, Tim Salimans, David Fleet, \& Mohammad Norouzi. Palette: Image-to-image diffusion models. In ACM SIGGRAPH 2022 Conference Proceedings, pp. 1–10, 2022a.

\hypertarget{ref49}{[49]} Ma, Xin, Wang, Yaohui, Jia, Gengyun, Chen, Xinyuan, Liu, Ziwei, Li, Yuan-Fang, Chen, Cunjian, \& Qiao, Yu. "Latte: Latent Diffusion Transformer for Video Generation." arXiv preprint arXiv:2401.03048, 2024.

\hypertarget{ref50}{[50]} Peebles, William, \& Xie, Saining. "Scalable Diffusion Models with Transformers." arXiv preprint arXiv:2212.09748, 2022.

\hypertarget{ref51}{[51]} Bao, Fan, Nie, Shen, Xue, Kaiwen, Cao, Yue, Li, Chongxuan, Su, Hang, \& Zhu, Jun. "All are Worth Words: A ViT Backbone for Diffusion Models." Proceedings of the IEEE/CVF Conference on Computer Vision and Pattern Recognition (CVPR), 2023.

\hypertarget{ref52}{[52]} Dosovitskiy, Alexey, Beyer, Lucas, Kolesnikov, Alexander, Weissenborn, Dirk, Zhai, Xiaohua, Unterthiner, Thomas, Dehghani, Mostafa, Minderer, Matthias, Heigold, Georg, Gelly, Sylvain, Uszkoreit, Jakob, \& Houlsby, Neil. "An Image is Worth 16x16 Words: Transformers for Image Recognition at Scale." arXiv:2010.11929, 2021.

\hypertarget{ref53}{[53]} Perez, Ethan, Strub, Florian, de Vries, Harm, Dumoulin, Vincent, \& Courville, Aaron C. "FiLM: Visual Reasoning with a General Conditioning Layer." Proceedings of the AAAI Conference on Artificial Intelligence (AAAI), 2018.

\hypertarget{ref54}{[54]} Jascha Sohl-Dickstein, Eric Weiss, Niru Maheswaranathan, \& Surya Ganguli. Deep unsupervised learning using nonequilibrium thermodynamics. In International Conference on Machine Learning, pp. 2256–2265. PMLR, 2015.

\hypertarget{ref55}{[55]} Dao, Tri, Fu, Daniel Y., Ermon, Stefano, Rudra, Atri, \& Ré, Christopher. "FlashAttention: Fast and Memory-Efficient Exact Attention with IO-Awareness." Advances in Neural Information Processing Systems (NeurIPS), 2022.

\hypertarget{ref56}{[56]} Dao, Tri. "FlashAttention-2: Faster Attention with Better Parallelism and Work Partitioning." 2023.

\hypertarget{ref57}{[57]} Zhou, Yucheng, Li, Xiang, Wang, Qianning, \& Shen, Jianbing. "Visual In-Context Learning for Large Vision-Language Models." arXiv:2402.11574, 2024.

\hypertarget{ref58}{[58]} Meng, Lingchen, Lan, Shiyi, Li, Hengduo, Alvarez, Jose M., Wu, Zuxuan, \& Jiang, Yu-Gang. "SEGIC: Unleashing the Emergent Correspondence for In-Context Segmentation." arXiv:2311.14671, 2023.

\hypertarget{ref59}{[59]} Julieta Martinez, Michael J Black, \& Javier Romero. On human motion prediction using recurrent neural networks. In Proceedings of the IEEE conference on computer vision and pattern recognition, pp. 2891–2900, 2017.

\hypertarget{ref60}{[60]} Romero, Javier, Tzionas, Dimitrios, \& Black, Michael J. "Embodied Hands: Modeling and Capturing Hands and Bodies Together." ACM Transactions on Graphics (Proc. SIGGRAPH Asia), ACM, November 2017.

\hypertarget{ref61}{[61]} Radford, Alec, Wu, Jeff, Child, Rewon, Luan, David, Amodei, Dario, \& Sutskever, Ilya. "Language Models are Unsupervised Multitask Learners." 2019.

\hypertarget{ref62}{[62]} Shafir, Yoni, Tevet, Guy, Kapon, Roy, \& Bermano, Amit Haim. "Human Motion Diffusion as a Generative Prior." ICLR 2024, 2024.

\hypertarget{ref63}{[63]} Lugmayr, Andreas, Danelljan, Martin, Gool, Luc Van, et al. "RePaint: Inpainting using Denoising Diffusion Probabilistic Models." Computer Vision and Pattern Recognition (CVPR), 2022.

\hypertarget{ref64}{[64]} Mu, Norman, Ji, Jingwei, Yang, Zhenpei, Harada, Nate, Tang, Haotian, Chen, Kan, Qi, Charles R., Ge, Runzhou, Goel, Kratarth, Yang, Zoey, Ettinger, Scott, Al-Rfou, Rami, Anguelov, Dragomir, \& Zhou, Yin. "MoST: Multi-modality Scene Tokenization for Motion Prediction." arXiv:2404.19531, 2024.

\hypertarget{ref65}{[65]} Jiang, Biao, Chen, Xin, Liu, Wen, Yu, Jingyi, Yu, Gang, \& Chen, Tao. "MotionGPT: Human Motion as a Foreign Language." Advances in Neural Information Processing Systems, vol. 36, 2024.

\hypertarget{ref66}{[66]} Chen, Xin, Jiang, Biao, Liu, Wen, Huang, Zilong, Fu, Bin, Chen, Tao, \& Yu, Gang. "Executing your Commands via Motion Diffusion in Latent Space." Proceedings of the IEEE/CVF Conference on Computer Vision and Pattern Recognition (CVPR), 2023, pp. 18000-18010.

\hypertarget{ref67}{[67]} Felix G Harvey \& Christopher Pal. Recurrent transition networks for character locomotion. In SIGGRAPH Asia 2018 Technical Briefs, pp. 1–4. 2018.

\hypertarget{ref68}{[68]} Manuel Kaufmann, Emre Aksan, Jie Song, Fabrizio Pece, Remo Ziegler, \& Otmar Hilliges. Convolutional autoencoders for human motion infilling. In 2020 International Conference on 3D Vision (3DV), pp. 918–927. IEEE, 2020.

\hypertarget{ref69}{[69]} Yinglin Duan, Tianyang Shi, Zhengxia Zou, Yenan Lin, Zhehui Qian, Bohan Zhang, \& Yi Yuan. Single-shot motion completion with transformer. arXiv preprint arXiv:2103.00776, 2021.

\hypertarget{ref70}{[70]} Mathis Petrovich, Michael J. Black, \& Gul Varol. Action-conditioned 3D human motion synthesis with transformer VAE. In International Conference on Computer Vision (ICCV), pp. 10985– 10995, October 2021.

\hypertarget{ref71}{[71]} Austin Patel, Andrew Wang, Ilija Radosavovic, \& Jitendra Malik. Learning to imitate object interactions from internet videos. arXiv preprint arXiv:2211.13225, 2022.

\hypertarget{ref72}{[72]} Xue Bin Peng, Angjoo Kanazawa, Jitendra Malik, Pieter Abbeel, \& Sergey Levine. Sfv: Reinforcement learning of physical skills from videos. ACM Transactions On Graphics (TOG), 37(6):1–14, 2018.

\hypertarget{ref73}{[73]} Vasileios Vasilopoulos, Georgios Pavlakos, Sean L Bowman, J Diego Caporale, Kostas Daniilidis, George J Pappas, \& Daniel E Koditschek. Reactive semantic planning in unexplored semantic environments using deep perceptual feedback. IEEE Robotics and Automation Letters, 5(3):4455– 4462, 2020.

\hypertarget{ref74}{[74]} Chung-Yi Weng, Brian Curless, Pratul P Srinivasan, Jonathan T Barron, \& Ira Kemelmacher-Shlizerman. HumanNeRF: Free-viewpoint rendering of moving people from monocular video. In CVPR, 2022.

\hypertarget{ref75}{[75]} Owen Pearl, Soyong Shin, Ashwin Godura, Sarah Bergbreiter, \& Eni Halilaj. Fusion of video and inertial sensing data via dynamic optimization of a biomechanical model. Journal of Biomechanics, 155:111617, 2023.

\hypertarget{ref76}{[76]} Angjoo Kanazawa, Michael J Black, David W Jacobs, \& Jitendra Malik. End-to-end recovery of human shape and pose. In CVPR, 2018.

\hypertarget{ref77}{[77]} Nikos Kolotouros, Georgios Pavlakos, Michael J Black, \& Kostas Daniilidis. Learning to reconstruct 3D human pose and shape via model-fitting in the loop. In ICCV, 2019.

\hypertarget{ref78}{[78]} Prafulla Dhariwal \& Alexander Nichol. Diffusion models beat gans on image synthesis. In NeurIPS, 2021.

\hypertarget{ref79}{[79]} Aditya Ramesh, Prafulla Dhariwal, Alex Nichol, Casey Chu, \& Mark Chen. Hierarchical textconditional image generation with clip latents. arXiv preprint arXiv:2204.06125, 2022.

\hypertarget{ref80}{[80]} Diederik Kingma \& Jimmy Ba. Adam: A method for stochastic optimization. In ICLR, 2015.

\hypertarget{ref81}{[81]} Ilya Loshchilov \& Frank Hutter. Decoupled weight decay regularization. arXiv:1711.05101, 2017.

}

\end{document}